\documentclass[final,3p,times,twocolumn,number]{elsarticle}

\usepackage{amssymb}
\usepackage{amsmath}

\usepackage{url}
\usepackage{float}
\usepackage{subcaption}
\usepackage{tabularx}
\usepackage{graphicx}
\usepackage{caption}

\journal{}

\begin{document}

\begin{frontmatter}


\title{BioNIC: Biologically Inspired Neural Network for Image Classification Using Connectomics Principles}
\author[aff1]{Diya Prasanth\corref{cor1}}
\ead{prasanthdiya0448@students.esuhsd.org}
\author[aff2]{Matthew Tivnan}
\ead{mtivnan@mgh.harvard.edu}

\cortext[cor1]{Corresponding author : prasanthdiya0448@students.esuhsd.org}
\affiliation[aff1]{organization={Accel Middle College},
            addressline={3095 Yerba Buena Road}, 
            city={San Jose},
            postcode={95135}, 
            state={CA},
            country={USA}}
\affiliation[aff2]{organization={Radiology Department},
            addressline={Harvard Medical School \& Massachusetts General Hospital}, 
            city={Boston},
            postcode={02114}, 
            state={MA},
            country={USA}}


\begin{abstract}
We present BioNIC, a multi-layer feedforward neural network for emotion classification, inspired by detailed synaptic connectivity graphs from the MICrONs dataset ~\cite{bae-2025, ding-2025}. At a structural level, we incorporate architectural constraints derived from a single cortical column of the mouse Primary Visual Cortex(V1): connectivity imposed via adjacency masks, laminar organization, and graded inhibition representing inhibitory neurons. At the functional level, we implement biologically inspired learning: Hebbian synaptic plasticity with homeostatic regulation, Layer Normalization, data augmentation to model exposure to natural variability in sensory input, and synaptic noise to model neural stochasticity. We also include convolutional layers for spatial processing, mimicking retinotopic mapping. The model performance is evaluated on the Facial Emotion Recognition task FER-2013~\cite{goodfellow-2013} and compared with a conventional baseline. Additionally, we investigate the impacts of each biological feature through a series of ablation experiments. While connectivity was limited to a single cortical column and biologically relevant connections, BioNIC achieved performance comparable to that of conventional models, with an accuracy of 59.77 ± 0.27\% on FER-2013. Our findings demonstrate that integrating constraints derived from connectomics is a computationally plausible approach to developing biologically inspired artificial intelligence systems. This work also highlights the potential of new generation peta-scale connectomics data in advancing both neuroscience modeling and artificial intelligence~\cite{Ngai-2025}.
\end{abstract}


\end{frontmatter}




\section{Introduction}
Emotion detection from facial expressions in images is a complex task due to subtle expression patterns, individual variability, variable pose, lighting, and occlusions. Widely used benchmarks in facial recognition research, such as FER-2013, have only about \(65\pm5\%\) human-level accuracy~\cite{goodfellow-2013, khaireddin-2021}. While proposed models for tasks like FER-2013 use the general principles of biological neural networks, the implementation of biologically faithful model architectures trained for these tasks remains largely unexplored, in part due to the lack of large-scale, high-resolution connectomics data. 

Recent advancements in electron microscopy and artificial intelligence have enabled large peta-scale connectomics datasets that can be used to reconstruct biological neural circuits~\cite{dorkenwald-2025}. The new datasets from projects such as FlyWire\cite{dorkenwald-2024, schlegel-2024}, FANC\cite{azevedo-2024}, HCP\cite{elam-2021} and MICrONS~\cite{bae-2025} provide an opportunity to explore novel approaches to designing artificial neural networks directly inspired by biological structures. Our work investigates a model architecture that reflects real mouse cortical connectivity and highlights the potential of biological organization to guide the design of expressive artificial neural network architectures for tasks like emotion detection.

Leveraging the MICrONs dataset, we created a neural network that incorporates the structural organization of the mouse visual cortex. We encode features observed in the visual cortex, such as cortical layering, inter- and intra-layer connections, and graded inhibition mechanisms. 

In the mouse visual system, the primary visual cortex(V1) performs initial processing of visual input\cite{busse-2018}. Higher visual areas, such as the insular cortex and amygdala, are involved in interpreting information. Our model does not include mapping of higher-order brain regions. The model processes crucial visual signals, which serve as the basis for visual patterns used to recognize emotions, and a final classification layer performs the functions of higher-order brain regions.

Our approach goes beyond working with conventional data-driven neural networks and engages in cross-disciplinary research in artificial intelligence(AI) and neuroscience. The model can serve as a platform to explore how changes in connectivity and inhibition mechanisms affect neural processing. Through ablation analysis we identify the biological features that are most important for the functionality and performance of the model.

\section{Related Work}

\subsection{FER-2013}
Several models have been proposed for the FER-2013 emotion classification task. The ICML 2013 winning solution~\cite{tang-2013} implemented a CNN~\cite{krizhevsky-2017} feature extractor with a linear SVM classifier, instead of the commonly used softmax, yielding a public validation score of 71.2\%. Subsequent works demonstrated that deeper CNNs with inception-style layers~\cite{mollahosseini-2022} and ensemble methods can  further improve accuracy~\cite{pramerdorfer-2016, khanzada-2020}. Parameter tuning and advanced optimization strategies with VGGNet achieved single model accuracies as high as 73.28\%~\cite{simonyan-2022, khaireddin-2021}. 

\subsection{Connectomics-inspired model architectures}

Studies have utilized structural priors derived from connectomics to inspire model architectures for artificial neural networks. Goulas et al. explored bio-inspired RNNs constrained by real neuroanatomical structures~\cite{goulas-2021}, while MouseNet{~\cite{shi-2022}} proposed a 2.1M parameter CNN model using the Allen Brain Observatory Visual Coding dataset. Another connectome-constrained task-optimized neural network study shows promising results for predicting the neural activity underlying visual motion detection ~ {~\cite{lappalainen-2024}}. Roberts et al. demonstrated deep neural network architectures based on topological wiring of \textit{C. Elegans} and mouse visual cortex connectomes{~\cite{roberts-2019}}. Bardozzo et al.~\cite{bardozzo-2024} proposed neural network models that use the connectome topology of \textit{C. Elegans} to construct deep and reservoir architectures which are more efficient than conventional neural networks. Damicelli et al. further examined the role of real brain connectome topologies in neural reservoir computing{~\cite{damicelli-2022}}. 

\section{Methods}

We performed all experiments on the FER-2013 facial emotion dataset created by Pierre Luc Carrier and Aaron Courville for the \textit{Challenges in Representation Learning: Facial Expression Recognition Challenge} at the 30th International Conference on Machine Learning(ICML) in 2013. BioNIC leverages publicly available connectomics data from the MICrONs project to model network topology inspired by biological connectomes.  

\subsection{FER-2013 Emotion Detection Dataset}
FER-2013 consists of 35,887 grayscale 48$\times$48 pixel images of various facial expressions. Images in the dataset are classified into seven emotion categories — happy, neutral, sad, angry, surprised, disgusted, and fearful — based on facial expression. The dataset consists of 28,709 training images and 7,178 test images, and the objective is to develop a model capable of accurately recognizing the emotion expressed in human facial photographs.

\begin{figure}[htbp]
    \centering
    \includegraphics[width=0.90\linewidth]{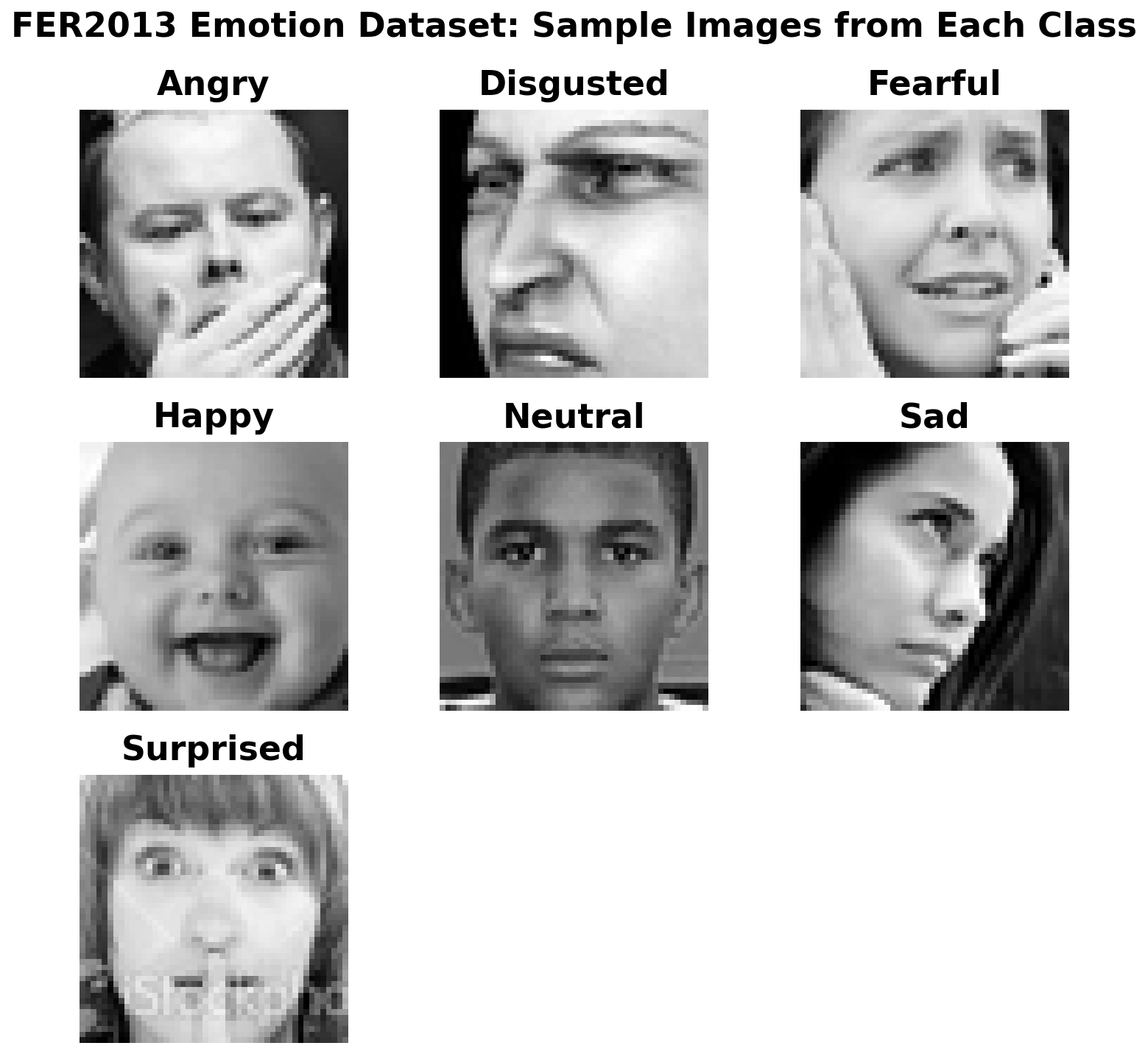}
    \caption{Sample images from the FER-2013 dataset}
    \label{fig:sample-figure}
\end{figure}

The challenges with the FER-2013 dataset include class imbalance from uneven sample sizes across categories and limited facial detail in low-resolution images. There is a significant degree of variation within each emotion class(intra-class), as well as overlap and similarity between some classes(inter-class)

\subsection{Connectomics Data}

The Machine Intelligence from Cortical Networks(MICrONS) is a large-scale IARPA project in collaboration with research teams at the Allen Institute, Baylor College of Medicine, and Princeton, aiming to create a dense reconstruction of the structural connections and functions of a millimeter volume of mouse visual cortex. The MICrONS dataset spans a section of mouse visual cortex and surrounding areas(V1, AL, RL, and LM) with approximately 200,000 cells and 523 million connections within the reconstructed volume. It combines in vivo imaging data with connectomic reconstructions to provide structural and functional data, enabling detailed mapping of neural circuitry and full 3D reconstruction of neurons, axons, dendrites, and synapses. Within the MICrONS dataset, the "minnie65\_public" data stack is the MICrONS project's proofread segmentation and annotation data, which reflects the mouse cortical connectome.

We used the MICrONs data access interface CAVE~\cite{dorkenwald-2025, caveclient-author-no-date} to retrieve data from  "minnie65\_public" data stack to define the models~\cite{bodor-2025}. 

\subsubsection{Cell Information}
CAVE client's~\cite{dorkenwald-2025} materialized view \textit{aibs\_cell\_info} provides the following cell information:
\begin{itemize}
  \item cell identifier
  \item broad type
  \item cell type
  \item spatial position
  \item cell\_type\_source
\end{itemize}
A subset of these cells identified by the \textit{cell\_type\_source} \textit{ allen\_v1\_column\_types\_slanted\_ref}  represents a single 10 $\mu$m cortical column in the V1 region. BioNIC aims to model the connectivity patterns in this selected cortical column. 

In the cell information above, cells are manually classified into non-neuronal, excitatory, and inhibitory classes~\cite{mao-2024} represented by the \textit{broad type} in \textit{aibs\_cell\_info}. They are further categorized into laminar and cell subtypes as indicated by \textit{cell type}~\cite{mao-2024}. 

\begin{table}[htbp]
  \centering
  \caption{Neurons per Cortical Cell Type}
  \scriptsize
  \begin{tabular}{|l|l|l|}
    \hline
    Cell Type & Number of Neurons & Type \\
    \hline
    23P & 349 & excitatory \\
    4P & 266 & excitatory \\
    5P-ET & 38 & excitatory \\
    5P-IT & 137 & excitatory \\
    5P-NP & 10 & excitatory \\
    6P-CT & 143 & excitatory \\
    6P-IT & 192 & excitatory \\
    6P-U & 28 & excitatory \\
    BC & 59 & inhibitory \\
    BPC & 33 & inhibitory \\
    MC & 41 & inhibitory \\
    NGC & 17 & inhibitory \\
    WM-P & 20 & excitatory \\
    \hline
  \end{tabular}
  \label{tab:celltypecounts}
\end{table}

\begin{figure}[htbp]
  \centering
  \begin{subfigure}[c]{0.45\textwidth}
    \centering
    \includegraphics[width=\linewidth]{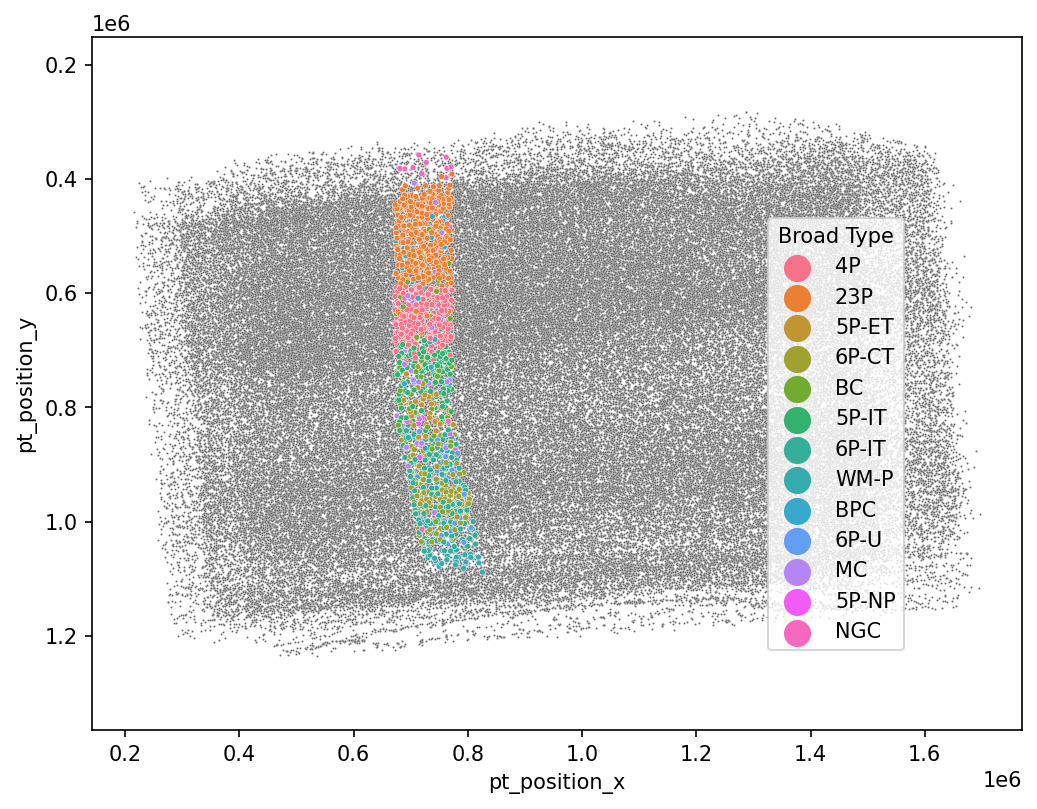}
    \caption{Nucleus detections highlighting one cortical column.}
    \label{fig:sample-figure1}
  \end{subfigure}
  \hfill
  \begin{subfigure}[c]{0.25\textwidth}
    \centering
    \includegraphics[width=\linewidth]{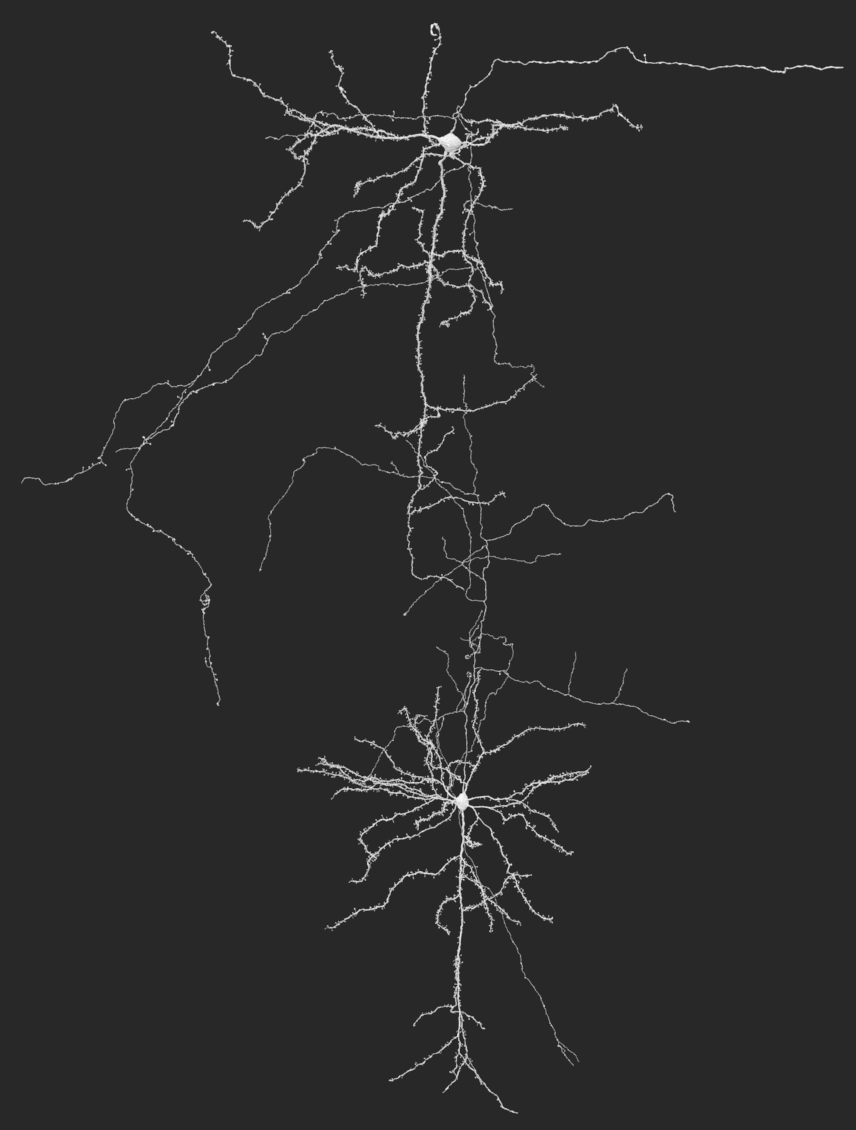}
    \caption{Visualization of two neurons from different cortical layers.}
    \label{fig:sample-figure2}
  \end{subfigure}
  \caption{(Left) Nucleus detections highlighting one cortical column; (Right) Visualization of two neurons from different cortical layers.}
  \label{fig:cortical-pair}
\end{figure}

\subsubsection{Synapse Information}
 From the synapse data, two adjacency matrices are constructed for the neurons in the selected cortical column: one representing the count of synapses between neurons(\textit{count\_adjacency}) and the other representing the sum of synapse sizes(\textit{synapsesize\_adjacency})~\cite{schneider-mizell-2025}.

The cell information and synapse information collectively establish full connectivity topology of the selected cortical column.

\subsection{Preprocessing and Data Augmentation}
During training, a randomized pipeline is applied to each image including grayscale conversion, random resized cropping(scale=0.80 to 1.0), rotation($\pm$15), horizontal flipping, and brightness/contrast jitter. We use this approach to mimic natural variability and regularize feature learning~\cite{shorten-2019}.  

\subsection{Model Architecture}

\subsubsection{BioNIC: Biologically Inspired Feedforward Neural Network}

\begin{figure*}
  \centering
  \includegraphics[width=\textwidth]{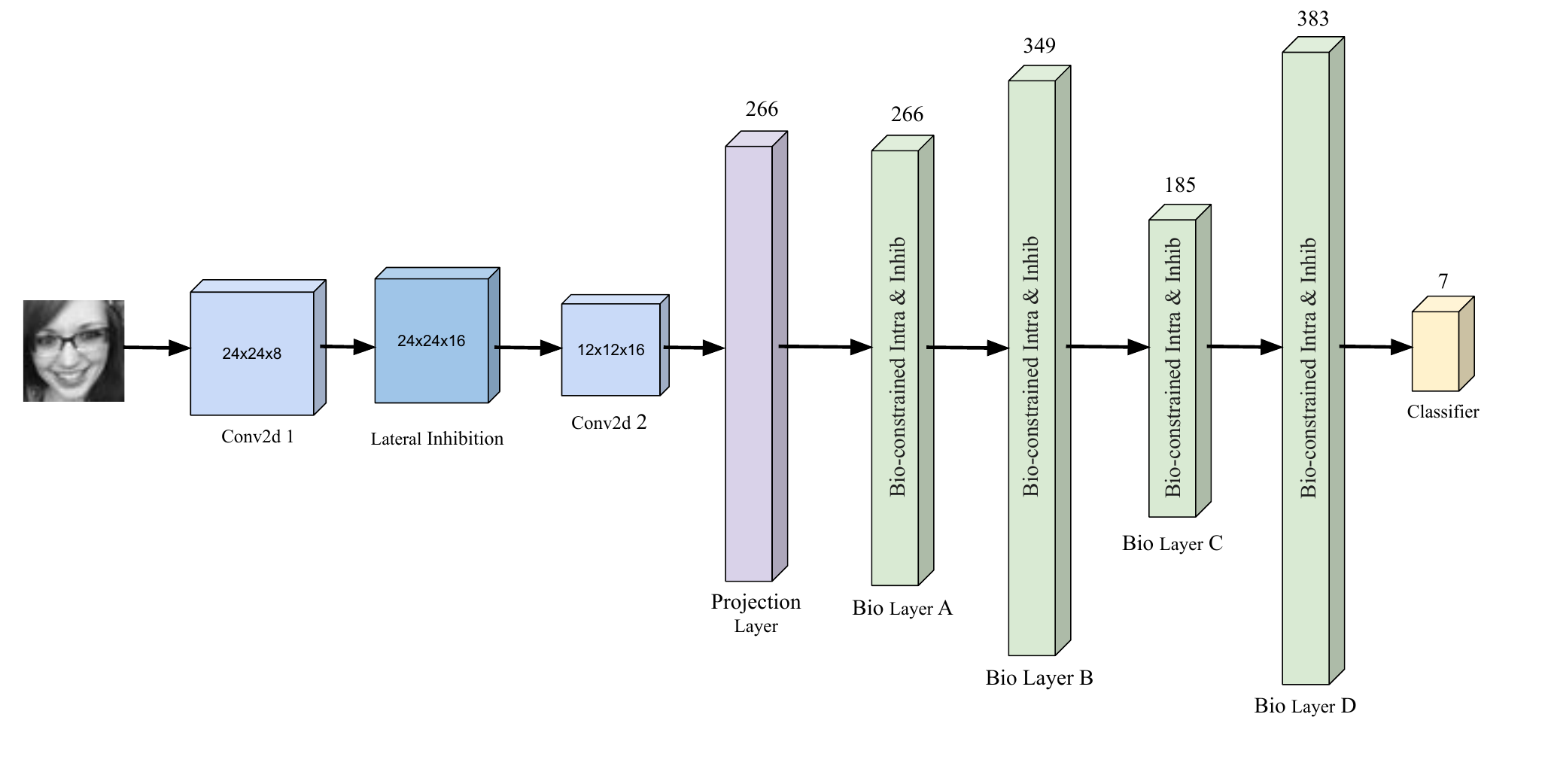}
  \caption{Model with biologically plausible inter-layer and intra-layer connections, graded inhibition, convolutional layers, and regularization}
  \label{fig:wide-image}
\end{figure*}
In the mouse visual system, neural activity starts with specialized photo-receptors in the retina that convert light to neural signals, extract features from the visual input, and encode that information~\cite{busse-2018} into parallel output channels. The dorsal Lateral Geniculate Nucleus(dLGN) relays this to V1 by retinotopic mapping~\cite{wandell-1995}, which maintains the spatial map of the visual input. V1 includes excitatory pyramidal neurons and inhibitory inter-neurons and has a six-layer architecture.

In a simplified view of V1, the information can be seen entering through layer 4, passing through superficial layers 2/3, deep layers 5/6, and then exiting the cortex to other areas \cite{olivas-2012}.

We classify neurons into hypothesized biological layers based on the simplified information flow and cell type. The size of each hypothesized layer (number of neurons) in the neural network is the aggregated neuronal count across all cell types within that layer.

\begin{figure}[htbp]
  \centering
  \begin{minipage}[t]{0.48\textwidth}
    \centering
    \includegraphics[height=0.8\linewidth]{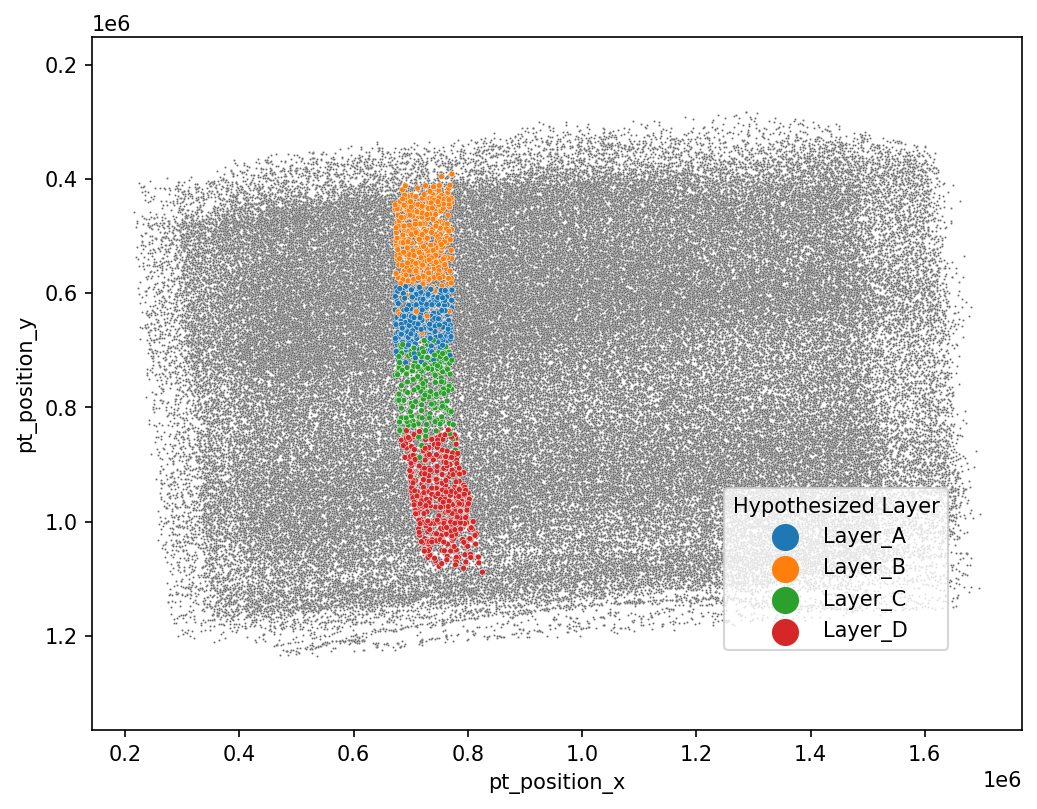}
    \caption{Hypothesized layers in one cortical column}
    \label{fig:sample-figure3}
  \end{minipage}%
  \hspace{0.02\textwidth}
  \begin{minipage}[t]{0.48\textwidth}
    \centering
    \renewcommand{\arraystretch}{1.3}
    \scriptsize
    \begin{tabularx}{\linewidth}{|l|X|c|}
      \hline
      Layer & Cell Types & Neurons \\
      \hline
      Layer A & 4P & 266 \\
      Layer B & 23P & 349 \\
      Layer C & 5P-ET, 5P-IT, 5P-NP & 185 \\
      Layer D & 6P-CT, 6P-IT, 6P-U, WM-P & 383 \\
      \hline
    \end{tabularx}
    \captionof{table}{Hypothesized Biological Layers and their Cell Types}
  \end{minipage}
\end{figure}

The model architecture includes a sequence of hypothesized biological layers(Layer A through Layer D), and connectivity between these layers is computed from the biological adjacency matrix. This empirical connectivity matrix constrains each layer’s weight matrix by applying an element-wise mask—preserving only those synaptic connections that are actually observed in the biological adjacency matrix.

The model takes 48×48 grayscale images as input. Two sequential convolutional layers, each followed by a ReLU activation~\cite{nair-2010}, project the image features to the size of the first biologically-inspired layer. The convolutional layers process the spatial input and transform it into a feature vector of the same size as the first biologically-inspired layer, preparing the data for biologically motivated computation downstream. Subsequent hidden layers correspond to the defined biological layers. The connections between these layers strictly follow the calculated biological connectivity. The final linear classifier layer maps the output of the last biological layer to the 7 emotion classes. 

The model uses Cross Entropy Loss, a commonly used loss function for classification tasks. The Adam~\cite{kingma-2014} optimizer updates the model's parameters during training.

Dense intra-layer connections enable local extraction, and inter-layer connections enable hierarchical processing. Synaptic connections preserve spatial relationships in the visual scene during processing. Stochasticity introduced by the noise in synaptic transmission and neuronal firing enhances generalization. Interneurons apply graded inhibition to other neurons to regulate their firing. They regulate how other neurons respond to specific features of the input and prevent runaway excitation. 

\subsubsection{BioNIC: Full model}
\begin{enumerate}
\item{Convolutional Layers: Input data is processed by two convolutional layers that act as a two-stage feature extractor. The first layer transforms inputs using 8 learnable filters. A lateral inhibition layer follows and suppresses neurons based on the activity of their spatial neighbors.

The lateral inhibition layer mimics inhibitory interactions in early visual pathways, such as those in the retina and thalamic relay(dLGN). In the mouse visual system, these inhibitory interactions enhance spatial differentiation. In these stages of real vision, neighboring neurons suppress each other’s activity through local connections, sharpening the signal that is ultimately delivered to cortical Layer 4.

The second convolutional layer receives the output from the lateral inhibition layer, applies 16 new filters, and uses ReLU for nonlinear transformation. The input features are then projected to the size of the first hypothesized layer.
$$h_1 =\mathrm{Conv}(x, W_1) + b_1, $$
where:
\begin{itemize}
    \item $x$ is the input tensor
    \item $W_1$ is the set of learnable weights
    \item $b_1$ is the bias 
\end{itemize}

$$h_1^{inh} = h_1 - \alpha \cdot \mathrm{AvgPool}(h_1), $$
where:
\begin{itemize}
    \item where $\alpha$ is the inhibition strength
    \item AvgPool occurs over a local spatial neighborhood
\end{itemize}

$$h_1^{act} =\mathrm{ReLU}(h_1^{inh}) $$
$$h_{conv} = \mathrm{ReLU}(\mathrm{Conv}(h_1^{act}, W_2) + b_2),$$
where:
\begin{itemize}
    \item $W_2$ is the set of learnable weights
    \item $b_2$ is the bias
\end{itemize}

\item{Hierarchical Attention:
The model implements hierarchical attention including spatial and channel attention mechanisms based on the design principles of \textit{CBAM}~\cite{woo-2018}} to learn which feature types and spatial regions matter.  Channel attention uses a bottleneck architecture~\cite{he-2016} following the \textit{Squeeze-and-Excitation} approach~\cite{hu-2018}. The channel attention compresses 16 input channels to 2 and expands back to 16.

$$z=\mathrm{GAP}(x)$$
$$s = \sigma(W_2(\mathrm{ReLU}(W_1(z))))$$
$$x_{ca} = x \odot s$$
where:
\begin{itemize}
    \item GAP denotes the global average pooling function
    \item W1 and W2 are learned weights
    \item $\sigma$ is the sigmoid activation function
\end{itemize}

Spatial attention uses 3x3 convolutions.
and spatial attention is based on the design principles of \textit{CBAM}~\cite{woo-2018}}. 

$$M_s = \sigma(\mathrm{Conv}_{3\mathrm{x}3}(x_{ca}))$$
$$x_{out}=x_{ca} \odot M_s$$

The flattened output is then passed to biological layers.

$$x_{flat}=\mathrm{Flatten}(x_{out})$$

\item{Biological Layers: For each $k$ from 1 to $L-1$, the input is the previous layer's output, $h^{inter}_{k-1}$. Let $n_{k-1}$ be the size of layer $k-1$ and $n_k$ be the size of layer $k$. The inter-layer connections are masked. 
$$h_k^{inter} = \mathrm{ReLU}((W_k \odot M_k)h^{inter}_{k-1}+b_k)$$
where:
\begin{itemize}
  \item $W_k$ is the full weight matrix connecting layers $k-1$ and $k$ ( $n_k$ x $n_{k-1}$)
  \item $M_k$ is the mask derived from the adjacency matrix for the connections between cell type in layer $k-1$ and $k$ (size $n_k$ x $n_{k-1}$)
  \item $b_k$ is the bias vector for layer $k$(size $n_k$)
\end{itemize}}

\item{Intra-layer connections:
In addition to inter-layer connections, the mouse visual cortex also includes intra-layer connections between neurons. Intra-layer connections are also implemented as linear layers and constrained by masks derived from the biological connectivity data, enabling neurons within a layer to influence one another. The output of layer k is obtained by summing the inter-layer and intra-layer inputs and applying the activation function.
$$h_k^{intra} = (U_k \odot N_k)h_k^{inter} + c_k $$
$$h_k^{comb}  = h_k^{inter} + h_k^{intra}$$
where:
\begin{itemize}
    \item $U_k$ is the intra-layer weight matrix
    \item $N_k$ is the intra-layer mask
    \item $c_K$ is the bias
\end{itemize}}

\item{Graded inhibition:
The previous sections included the functionality of excitatory neurons but did not include that of inhibitory neurons. We incorporate a biologically realistic form of inhibition that scales neuronal activation based on the number of incoming inhibitory connections, as determined by biological data. We assume the neurons with more incoming inhibitory connections are suppressed.
To incorporate inhibitory neuron functionality into the model, we introduce an inhibitory mask and an inhibitory scaling factor. If a neuron in a layer receives an inhibitory input, we set the corresponding value in the mask to 1; otherwise, we set it to 0. The scaling factor is a value between 0 and 1, and can be tuned.

In each layer, after calculating the weighted sum, we apply the inhibitory scaling factor specified by the mask, as shown below. This should have the effect of receiving an input from an inhibitory neuron. 

$$s^{(k)} = 1-\alpha\cdot \frac{I_k}{max(I_k)+\epsilon}$$
where:
\begin{itemize}
    \item $I_k$ is the vector(size $n_k$) of total incoming inhibitory connection counts for each neuron in layer $k$
    \item $\alpha$ is a learnable inhibitory scaling factor. Inhibitory scaling for each neuron is based on its normalized inhibitory connection count.
    \item $\epsilon$ is a small value to prevent division by zero
    \item ${s}^{(k)}$ is a vector scaling factor for layer $k$
\end{itemize}

The inhibited output is obtained by element-wise multiplication of the combined output and inhibition scaling:
$$\tilde{h}^{(k)}=h_{comb}^{(k)} \odot s^{(k)}$$
The output of layer $k$ is obtained by applying the activation function to the inhibited output: 
$$h_k = \mathrm{ReLU}(\tilde{h}^{(k)})$$}

\item{Data Augmentation:
When we train the model on images, the lack of variability in the training dataset makes it difficult to improve the model's ability to generalize ~\cite{suto-2024}. In real-world scenarios, biological neural networks will be able to process visual imagery with varying lighting, angles, visibility, etc. We achieve this by using data augmentation techniques: grayscale conversion, random resized cropping(scale=0.8 to 1.0), rotation($\pm$15), horizontal flipping(probability=0.5), and brightness/contrast jitter(brightness=0.15, contrast=0.15). We train the model on augmented data; the test dataset remains unchanged.}   

\item{Weight Decay~\cite{krogh-1991}:
Biological neural networks have natural mechanisms that impose constraints, such as synaptic efficiency and physical resource limitations, which improve generalization. In artificial neural networks, the mechanism for achieving this is weight decay. We enable weight decay by adding L2 regularization to the optimizer to help with overfitting and improve generalization.}

\item{Synaptic noise:
Synaptic noise is introduced in the model by injecting Gaussian noise into the outputs of each biological layer during training. This noise is generated with a configurable standard deviation($\sigma=0.06$) and is added to the post-inhibition activations, only when the model is in training mode.
$$h_k^\mathrm{noise} = h_k + \eta_k$$

$$\eta_k \sim \mathcal{N}(0, \sigma_k^2)$$
where:
\begin{itemize}
\item $h_k$ is the pre-noise output of layer 
\item $k$ and $h_k^\mathrm{noise}$ is the output after adding synaptic noise
\item $\eta_k$ sampled element-wise from a normal distribution with mean $0$ and variance $\sigma_k^2$.
\end{itemize}}

\item{Layer Normalization:
We chose Layer Normalization~\cite{shen-2021} over batch normalization\cite{ioffe-2015} to maintain biological plausibility in the model. LayerNorm normalizes activations independently for each input sample mirroring biological neurons. LayerNorm is applied at three critical points in the model architecture: after convolutional feature extraction stage to bridge into the biological layers, after each biological layer and in the output classifier before final predictions. LayerNorm approximates neural homeostatic plasticity and divisive normalization\cite{shen-2021, lian-2023} mechanisms in biological neurons which regulate firing rates based on local activity patterns.} 

\item{LR Scheduler: BioNIC uses the ReduceLROnPlateau~\cite{ReduceLROnPlateau-2023} learning rate scheduler, which adaptively reduces the learning rate based on model performance. In biological neural systems, learning rates change naturally from high plasticity to low plasticity as the brain matures~\cite{hensch-2004}. Early stages prioritize rapid learning and significant structural changes, which reduce over time to focus on finer tuning~\cite{wiesel-1963}, the model parallels this mechanism.}

\item{Output Classifier Layer: The output of the last biological layer $h_{L-1}$(size $n_{L-1}$) is passed through a final linear layer to calculate scores for each class of emotion. Let $C$ be the number of emotion classes. 
$$z = W_{out}h_{L-1}+b_{out}$$
where $z$ represents the raw scores, $W_{out}$ is the weight matrix of size $C$x$n_{L-1}$ and $b_{out}$ is the bias vector of size $C$.

The final prediction is obtained by applying a softmax to the outputs to get the probabilities, and then taking the argmax. 
$$\hat{y}=\mathrm{argmax}(\mathrm{Softmax}(z))$$}

\item{Hebbian Synaptic Plasticity: The model incorporates a simplified version of Hebbian plasticity~\cite{morris-1999} with homeostatic regularization~\cite{turrigiano-2000, zenke-2013} in biological layers. We update synaptic weights based on correlation between pre- and post-synaptic activity instantaneously during the training process. In addition to this homeostatic regularization maintains stable firing rate.
$$\Delta w=\alpha_{hebb} \cdot \sigma_{pre} \cdot \sigma_{post}-\alpha_{home} \cdot (f_{avg}-f_{target})$$
where:
\begin{itemize}
    \item $\sigma$ is the activation function and normalized functions to [0,1]
    \item $\alpha_{hebb}=5*10^{-4}$, the Hebbian learning rate
    \item $\alpha_{home}=5*10^{-4}$ is the homeostatic learning rate
    \item $f_{avg}$ is the average post-synaptic firing rate
    \item $f_{target}=0.35$
\end{itemize}
}
\end{enumerate}

\subsection{Ablation Variants}

We performed systematic ablations to study the contribution of each biological element by removing inhibition, intra-layer connections, convolutional stages, synaptic noise, weight decay, data augmentation, connectivity masks, lateral inhibition layer, LR Scheduler, and label smoothing. For comparison, we train a matched standard CNN baseline model lacking any bio-inspired features. All the experiments were performed under identical settings to ensure objective assessment.

\subsection{Training Procedures and Evaluation Metrics}

Models were trained using Adam\cite{kingma-2014}, with learning rate of $6*10^{-4}$ and weight decay $1*10^{-4}$. Cross-entropy loss~\cite{szegedy-2016, muller-no-date} was computed with balanced class weighting and label smoothing ($\alpha=0.1$) for regularization, robustness, and to reduce overconfidence. Both class weighting and label smoothing have a strong biological foundation: Class weighting scales model weights to account for imbalances, closely mirroring adaptive gain control induced by homeostatic plasticity in the brain. Adaptive gain controls stabilize neural activities by adjusting synaptic strengths. Label smoothing, another complementary regularization mechanism, constrains the model to produce probability distributions, as opposed to discrete scores. This mechanism is analogous to how information is distributed across cortical populations.
 
The learning rate was adaptively scheduled ( ReduceLROnPlateau~\cite{ReduceLROnPlateau-2023} ). All models used a batch size of 32 and up to 400 epochs, with early stopping~\cite{prechelt-1998} triggered after 40 epochs without improvement. To ensure reproducibility and robustness, we trained each model three times with different random seed initializations. All other hyper parameters remained the same across runs. 

Model effectiveness is assessed using test-set accuracy, average cross entropy loss, F1-Score, Confusion Matrix, and Area Under Curve( AUC )~\cite{fawcett-2005}. 

\section{Results}

We evaluated our full biologically inspired neural network model and multiple ablated variants on the FER-2013 dataset. All models trained under identical conditions are compared against a standard CNN baseline. Confusion Matrix and ROC Curve from the last run is used for per-class analysis. The main findings are summarized below.

\subsection{BioNIC(full model) vs. Standard CNN}

\subsubsection{Observations}
\begin{itemize}
    \item Accuracy: BioNIC achieves almost identical accuracy(60\%) to the standard CNN baseline
    \item Macro and Weighted F1 Scores are almost identical
    \item "Disgusted" Class: Standard model achieves high precision(0.88), but low recall(0.26) indicating that the model only predicts "disgusted" when it is confident. For the biological model, recall(0.40), F1(0.48), lower precision(0.59) suggests the model is more willing to predict "disgusted" cases and may sometimes predict over-confidently. 
    \item "Fearful" Class: Biological model has higher recall and F1, and both have the same precision. 
    \item "Happy" and "Surprised" classes: Performance is similar for these majority classes. 
    \item "Neutral", "Sad", and "Angry" Classes: Performance is similar between the models. 
    \item Generalization: Macro Average F1 is slightly higher for the Biological Model(0.56 vs. 0.55), indicating small recall gains for minority classes. 
\end{itemize}
\subsubsection{Interpretation}
\begin{itemize}
    \item BioNIC improves recall and balance for minority classes "disgusted" and "fearful", potentially due to constraints and regularization in the biological model. This improvement comes at the cost of some precision. 
    \item Standard CNN is comparatively more conservative in its predictions for some of the classes. 
\end{itemize}

\begin{table*}[ht]
\centering
\scriptsize
\caption{Comparison of Classification Metrics: Standard CNN vs. Full Biological Model}
\begin{tabular}{|l|c|c|c|c|c|c|}
\hline
\textbf{Class} & \textbf{CNN Prec.} & \textbf{Bio Prec.} & \textbf{CNN Recall} & \textbf{Bio Recall} & \textbf{CNN F1} & \textbf{Bio F1} \\
\hline
angry      & 0.48  & 0.48  & 0.53  & 0.52  & 0.50  & 0.50 \\
disgusted  & 0.88  & 0.59  & 0.26  & 0.40  & 0.40  & 0.48 \\
fearful    & 0.48  & 0.48  & 0.33  & 0.38  & 0.39  & 0.42 \\
happy      & 0.77  & 0.78  & 0.83  & 0.81  & 0.80  & 0.80 \\
neutral    & 0.52  & 0.54  & 0.64  & 0.60  & 0.58  & 0.57 \\
sad        & 0.51  & 0.49  & 0.44  & 0.46  & 0.47  & 0.47 \\
surprised  & 0.70  & 0.71  & 0.77  & 0.72  & 0.73  & 0.72 \\
\hline
\textbf{macro avg}      & 0.62  & 0.58  & 0.54  & 0.56  & 0.55  & 0.56 \\
\textbf{weighted avg}   & 0.60  & 0.59  & 0.60  & 0.60  & 0.59  & 0.59 \\
\textbf{accuracy}   & 0.60  & 0.60  &   &   &   &  \\
\hline
\end{tabular}
\label{tab:cnn_bio_comparison}
\end{table*}

Confusion matrix and ROC curves are very similar for both models with minor differences for disgusted, fearful and neutral classes. Notably, "disgusted" in FER-2013 is often misclassified as "anger" due to overlapping facial features\cite{khaireddin-2021}. We do see this behavior in the confusion matrix. However, the misclassification of "disgust" and "fear" due to very small sample sizes is not reflected in the confusion matrix. We do see the confusion between "sad" and "fear" as expected. 

The full model achieved a test accuracy of 59.77 ± 0.27\% and a macro-averaged F1-score of 0.5652 ± 0.0015, outperforming the ablation variants except ablated weight Decay.

\begin{figure}[htbp]
  \centering

  \begin{minipage}{0.5\textwidth}
    \centering
    \includegraphics[width=\textwidth]{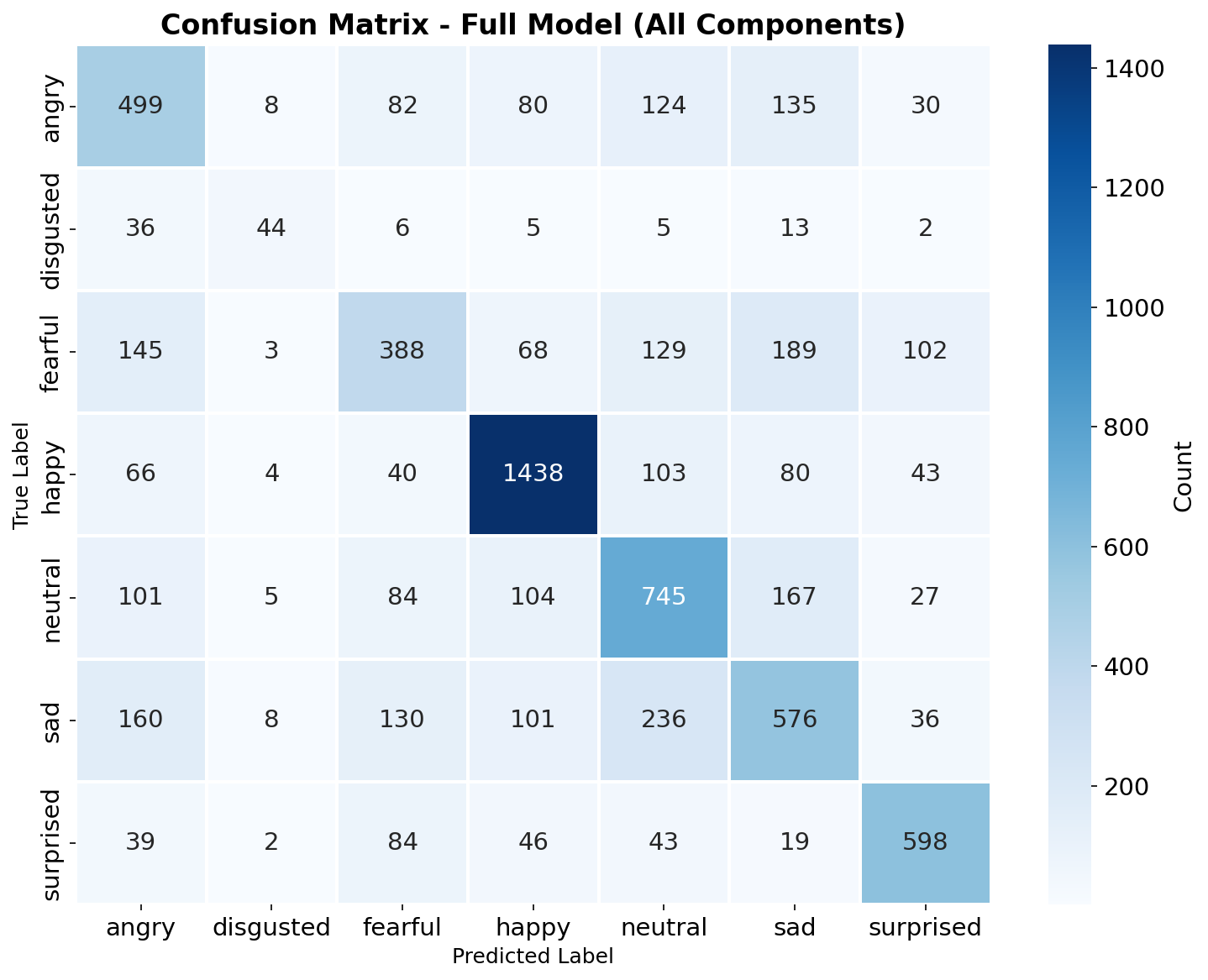}
    \caption*{Confusion Matrix}
  \end{minipage}%
  \hfill
  \begin{minipage}{0.5\textwidth}
    \centering
    \includegraphics[width=\textwidth]{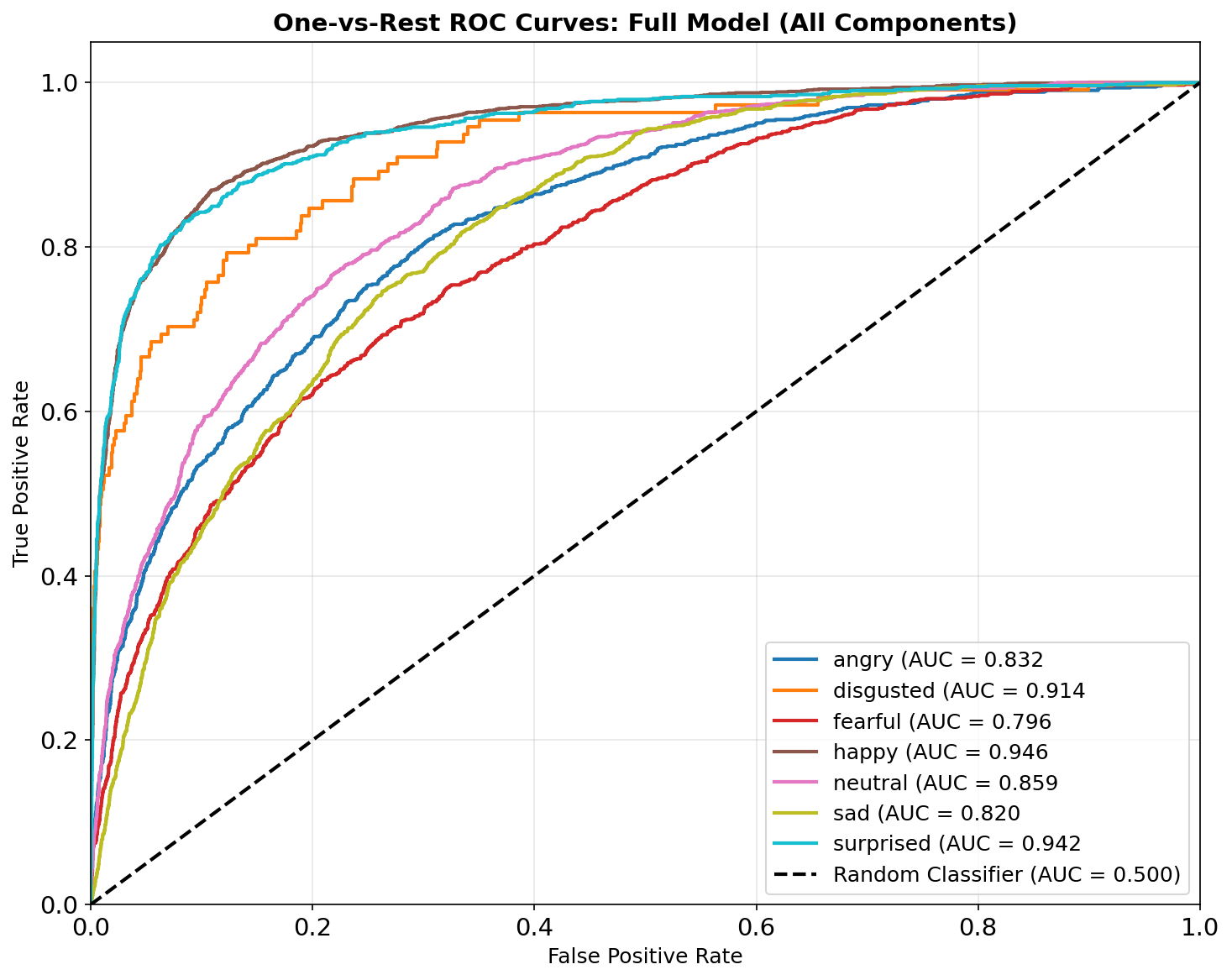}
    \caption*{ROC Curve}
  \end{minipage}

  \vspace{0.5em} 
\end{figure}
\begin{figure}[htbp]
  \centering

  \begin{minipage}{0.5\textwidth}
    \centering
    \includegraphics[width=\textwidth]{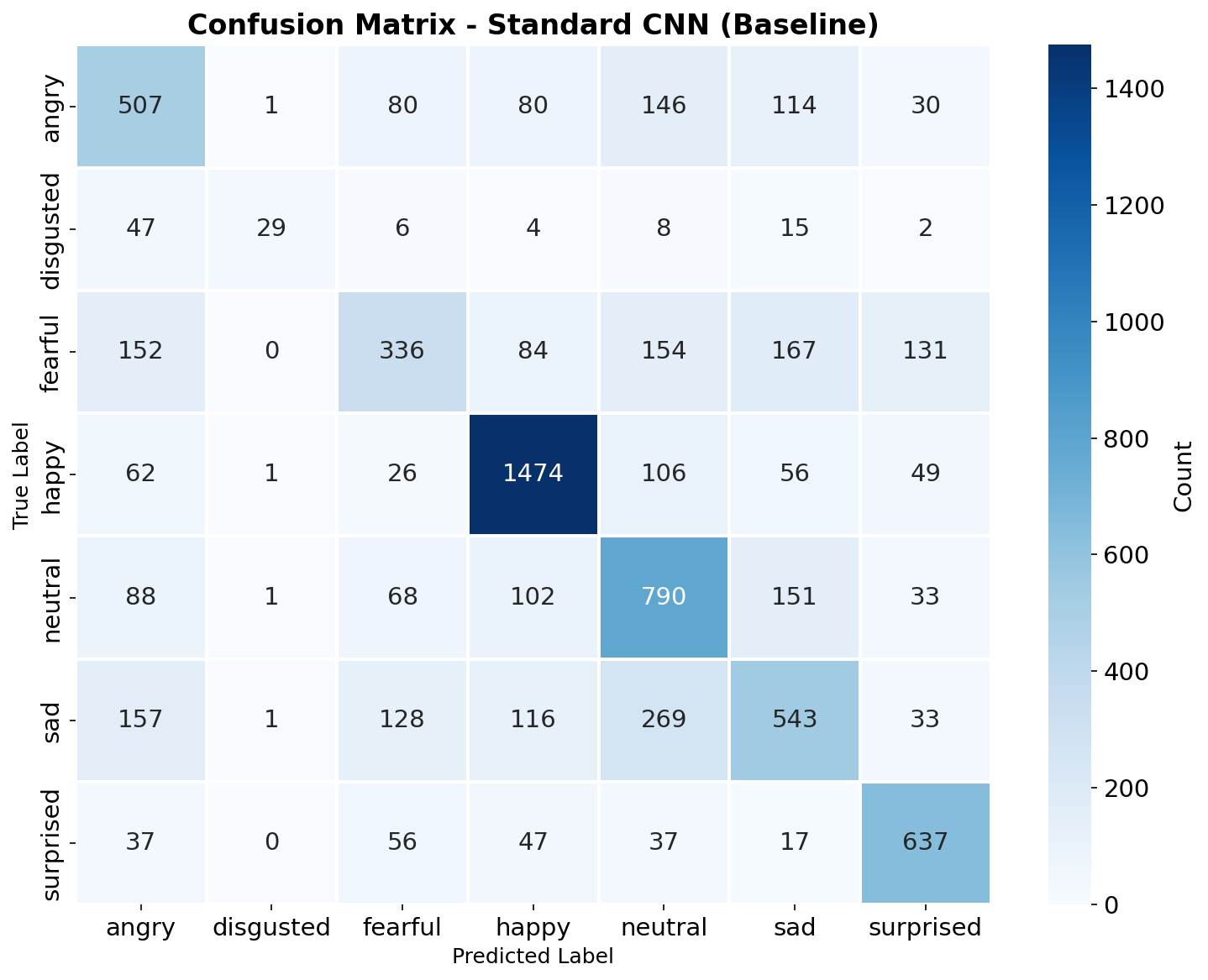}
    \caption*{Confusion Matrix}
  \end{minipage}%
  \hfill
  \begin{minipage}{0.5\textwidth}
    \centering
    \includegraphics[width=\textwidth]{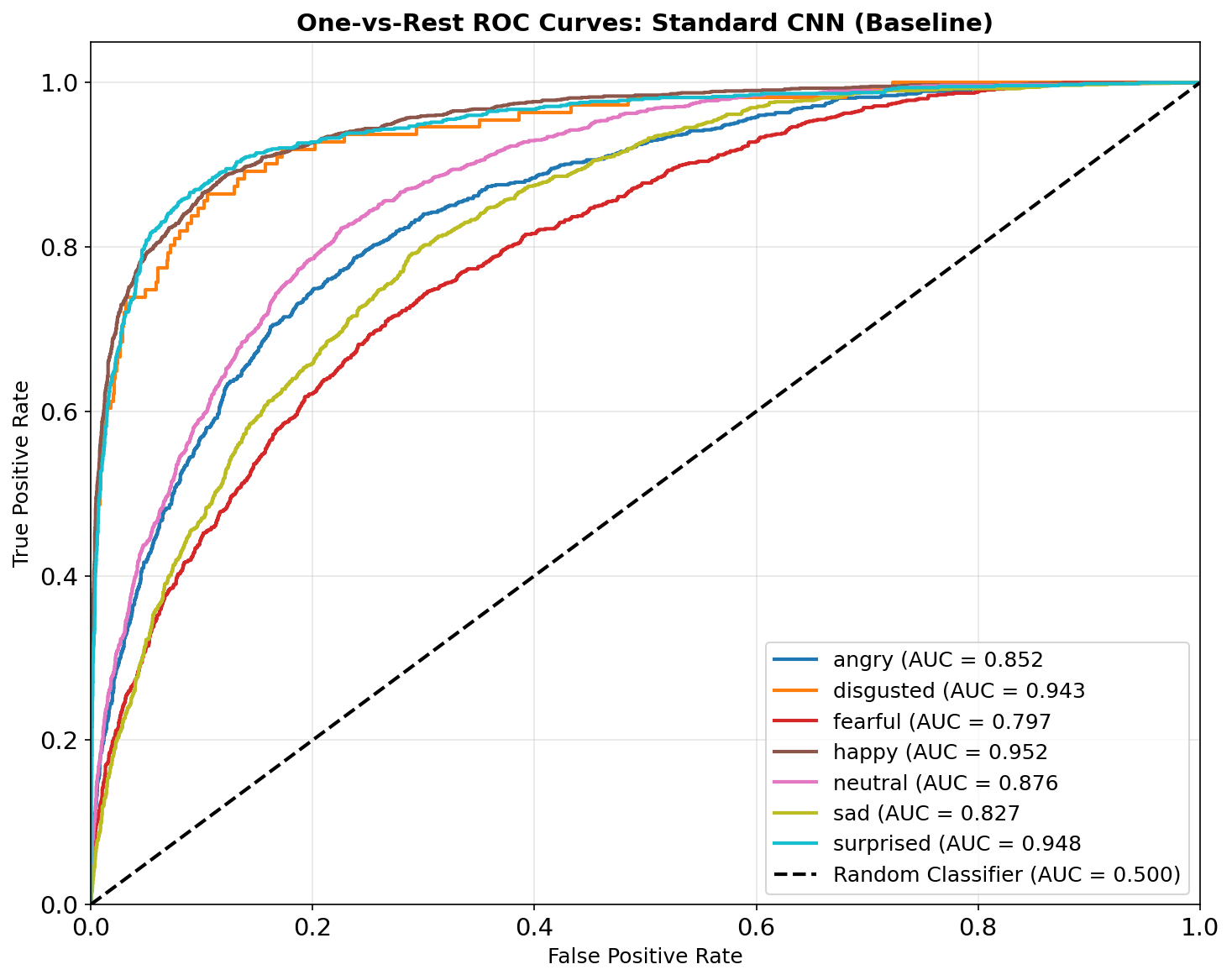}
    \caption*{ROC Curve}
  \end{minipage}
  \hfill
\end{figure}

\begin{figure}[htbp]
  \centering

  \begin{minipage}{0.49\textwidth}
    \centering
    \includegraphics[width=\textwidth]{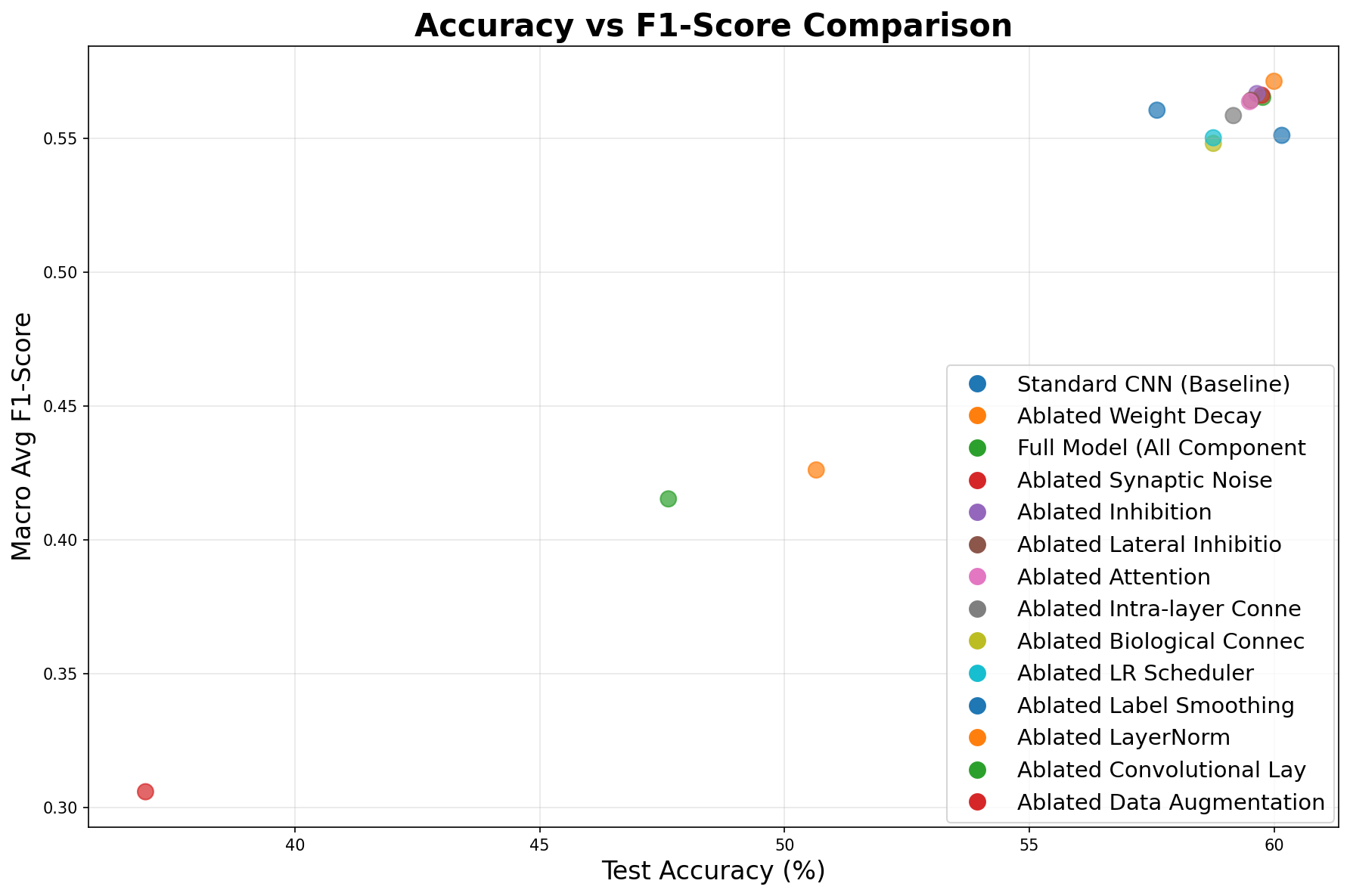}
    \caption*{Accuracy vs F1-Score}
  \end{minipage}%

\vspace{2em} 

  \begin{minipage}{0.49\textwidth}
    \centering
    \includegraphics[width=\textwidth]{images/AblationStudy.png}
    \caption*{Ablation Study}
  \end{minipage}

  \vspace{2em} 

  \begin{minipage}{0.50\textwidth}
    \centering
    \includegraphics[width=\textwidth]{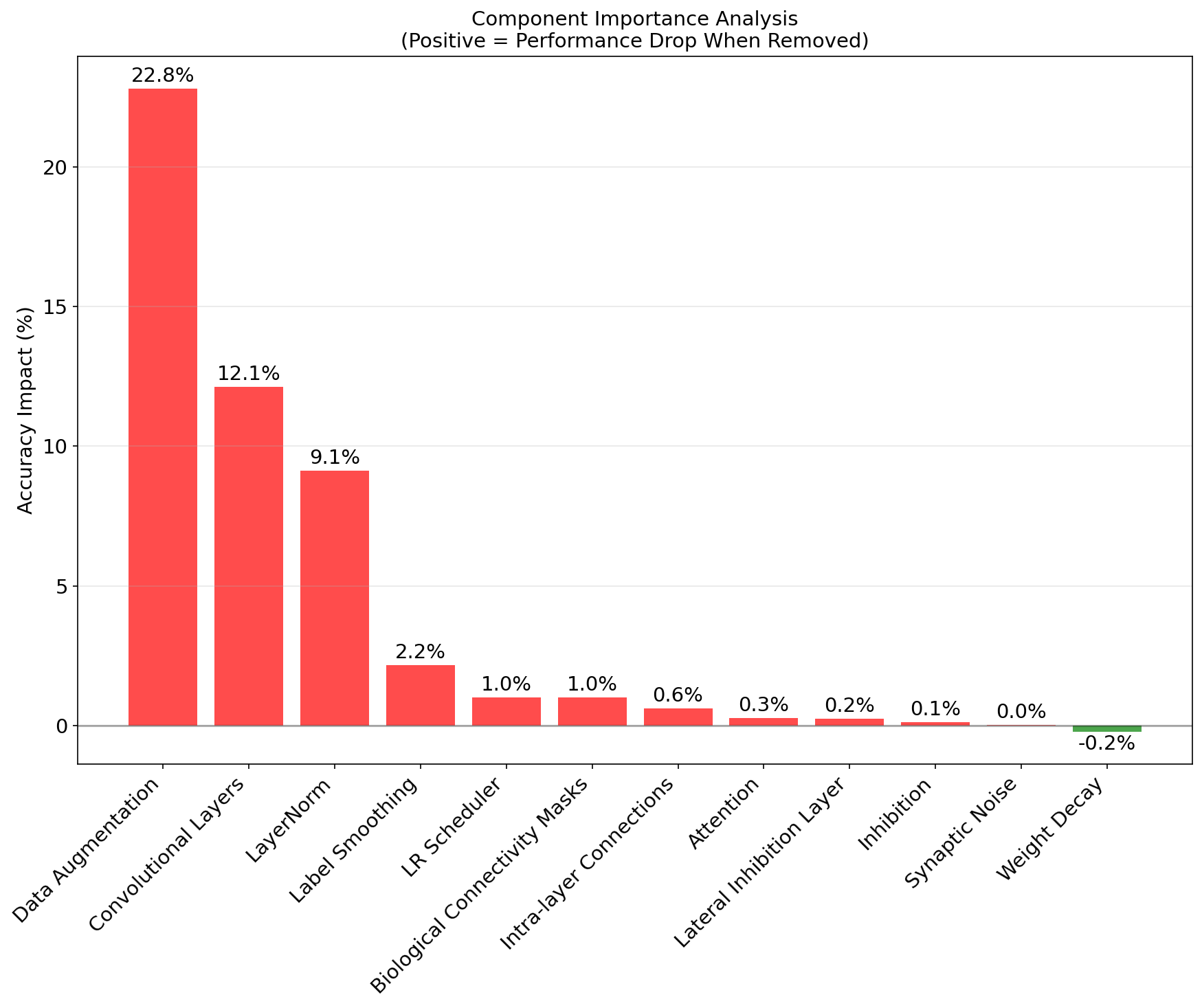}
    \caption*{Component Importance}
  \end{minipage}
\end{figure}

\subsection{Ablation studies}

Ablation experiments shows the functional importance of each biological element in model performance.

\begin{itemize}
    \item Ablated Weight Decay: Removing weight decay resulted in test accuracy of 60.00 ± 0.38, representing a slight improvement over the full model. Test performance improved across most classes, indicating that other regularization mechanisms and architectural constraints provided sufficient regularization.  
    \item Ablated Synaptic Noise: Eliminating synaptic noise did not affect the overall test accuracy significantly. The confusion matrix and ROC indicate minimal per class impact at the configured noise level.
    \item Ablated Inhibition: Removing the inhibition functionality resulted in a minimal performance degradation(roughly 0.5\% decrease in test accuracy). The model required more epochs to converge indicating that inhibition helped with faster convergence.
    \item Ablated Lateral Inhibition: Removing the lateral inhibition layer from the initial convolution layers led to a very small decrease of 0.24\% in the test accuracy with increased standard deviation indicating less stable predictions across runs.
    \item Ablated Attention: Removing both spatial and hierarchical attention mechanisms resulted in minimal performance drop of 0.27\% making this one of the least critical components in this study. This indicates that the spatial features convolutional layers already learned was sufficient and attention provided no significant benefit. However, the F1 variance was 4.4 times higher which suggests that attention did have some effect on regularization. 
    \item Ablated Intra-layer Connections: Removing the Intra-layer connections caused a minor reduction in accuracy. There was no evidence of class specific impacts suggesting that while intra-layer connectivity improved biological plausibility, it had minimal impact making it beneficial but not essential for this task. 
    \item Ablated Biological Connectivity Masks: While removing biological connectivity masks increased model capacity, this experiment yielded a minor decrease of 1.01\%, indicating that biological masks did not limit performance. Full biological model showed better class-wise accuracy for challenging-to-discriminate classes angry and fearful. This highlights that while biological connectivity masks improve class discrimination the absence of it doesn't prevent the rest of the architecture from capturing most of the discriminative capacity. 
    \item Ablated LR Scheduler: Removing the LR Scheduler resulted in slower convergence, more volatile test loss, less effective optimization later in training, and a drop in test accuracy to 58.76 ± 0.27. Inter-run variance remained similar to the full model, but varied across runs from 137 to 233 epochs before early stopping ended training. In later epochs, the model failed to fine tune weights  leading to suboptimal final performance indicating that the LR Scheduler played a role in achieving stabilization and peak test accuracy. 
    \item Ablated Label Smoothing: Removing label smoothing resulted in a noticeable performance decrease of roughly 2.2\%, to 57.61 ± 0.29. This indicates that label smoothing is one of the impactful regularization mechanisms for the model implementation. ROC analysis shows that AUC decreased for most classes indicating that label smoothing provided uniform regularization benefits across all classes. 
    \item Ablated Layer Normalization:
    Removing Layer normalization caused severe performance degradation with test accuracy dropping 9.12\% to 50.65 ± 1.85\% and macro F1 dropping 24.6\%. Training became unstable with 56\% increased training time and 3.3x higher variance. Convergence was unpredictable with 6.8 times accuracy variance, 22 times higher F1 variable across runs. This ablation confirms the importance of Layer normalization for stable optimization, making it 4th most critical component in this study.   
    
    \item Ablated Convolutional Layers: Removing convolutional layers resulted in a dramatic 12.14\% drop in test accuracy. All emotion classes were affected, with macro F1 dropping by 26.5\% and severely impacted convergence. This was the second largest drop in the ablation study, which highlights the importance of specially correlated feature extraction from the input images at the beginning of the processing pipeline. Losing the spatial connections makes robust recognition impossible.
    \item Ablated Data Augmentation: Removing all data augmentation led to a devastating 22.82\% in the test accuracy to 36.95 ± 1.34\% with macro F1 collapsing 45.9\%. All categories were impacted with the most underrepresented class disgust(-0.25 F1) and the most confused class fear(-0.16) impacted the worst, making ablated data augmentation the most critical component in the ablation study. The model reached early stopping 4.4x faster yet achieved worse test accuracy, indicating severe overfitting. This demonstrates the importance of exposure to rich varied inputs for learning to achieve generalization consistent with the neuro science principle\cite{raviv-2022, hernandez-garcia-2020}.
\end{itemize}

\begin{table*}[ht]
\centering
\scriptsize
\begin{tabular}{|l|c|c|c|c|c|r|}
\hline
Model Version & Test Accuracy \% & Test Loss & Train Accuracy \% & Train Loss & F1 Score & Parameters \\
\hline
Standard CNN (Baseline)           & 60.16 ± 0.45 & 0.7871 ± 0.0095 & 61.58 ± 0.74 & 0.7651 ± 0.0113 & 0.5510 ± 0.0061 & 1,381,223 \\
Abl. Weight Decay              & 60.00 ± 0.38 & 0.8478 ± 0.0135 & 70.41 ± 1.60 & 0.5937 ± 0.0290 & 0.5712 ± 0.0105 & 1,536,684 \\
Full Model (All Components)       & 59.77 ± 0.27 & 0.8348 ± 0.0047 & 68.63 ± 1.11 & 0.6272 ± 0.0231 & 0.5652 ± 0.0015 & 1,536,684\\
Abl. Synaptic Noise            & 59.75 ± 0.19 & 0.8372 ± 0.0083 & 68.23 ± 0.95 & 0.6324 ± 0.0176 & 0.5660 ± 0.0019 & 1,536,684 \\
Abl. Inhibition                & 59.65 ± 0.24 & 0.8409 ± 0.0052 & 69.38 ± 0.83 & 0.6137 ± 0.0152 & 0.5666 ± 0.0023 & 1,536,684 \\
Abl. Lateral Inhibition Layer  & 59.53 ± 0.42 & 0.8431 ± 0.0046 & 68.25 ± 1.47 & 0.6326 ± 0.0244 & 0.5641 ± 0.0056 & 1,536,684 \\
Abl. Attention                & 59.50 ± 0.22 & 0.8371 ± 0.0036 & 69.02 ± 0.33 & 0.6209 ± 0.0065 & 0.5636 ± 0.0066 & 1,536,684 \\
Abl. Intra-layer Connections   & 59.17 ± 0.15 & 0.8395 ± 0.0108 & 66.41 ± 1.57 & 0.6705 ± 0.0283 & 0.5584 ± 0.0036 & 1,536,684 \\
Abl. Connectivity Masks & 58.76 ± 0.57 & 0.8359 ± 0.0068 & 64.43 ± 2.38 & 0.7022 ± 0.0411 & 0.5480 ± 0.0104 & 1,536,684 \\
Abl. LR Scheduler             & 58.76 ± 0.27 & 0.8257 ± 0.0059 & 62.69 ± 1.19 & 0.7422 ± 0.0192 & 0.5501 ± 0.0074 & 1,536,684 \\
Abl. Label Smoothing          & 57.61 ± 0.29 & 1.1817 ± 0.0055 & 62.49 ± 0.73 & 0.9342 ± 0.0229 & 0.5604 ± 0.0006 & 1,536,684 \\
Abl. LayerNorm                & 50.65 ± 1.85 & 0.9481 ± 0.0359 & 49.53 ± 1.94 & 0.9583 ± 0.0352 & 0.4260 ± 0.0337 & 1,536,684 \\
Abl. Convolutional Layers     & 47.63 ± 0.76 & 1.0213 ± 0.0057 & 48.85 ± 1.03 & 0.9795 ± 0.0165 & 0.4152 ± 0.0135 & 1,535,308 \\
Abl. Data Augmentation        & 36.95 ± 1.34 & 1.2207 ± 0.0402 & 50.98 ± 2.89 & 20.4665 ± 13.7304 & 0.3058 ± 0.0092 & 1,536,684 \\
\hline
\end{tabular}
\caption{Comparison of Models with Test and Train Performance Metrics(masked parameters excluded)}
\label{tab:model_comparison}
\end{table*}


\section{Discussion}

The architecture shows strengths in identifying underrepresented classes and managing inter-class visual similarities. BioNIC improved recall by 14 percentage points(from 0.26 to 0.40) for the "disgusted" class, and by 5 percentage points for "fearful"(0.33 to 0.38). Precision rates declined for these visually similar classes, especially "disgust", indicating greater willingness to predict these classes.

\subsection{Structural vs. Functional Biological Constraints}
\begin{table}[ht]
\centering
\scriptsize
\caption{Effects of Structural Constraints}
\begin{tabular}{|l|c|}
\hline
Mechanism & Change in Accuracy\\
\hline
Connectivity Masks & +1.01\% \\
Intra-layer Connections & +0.60\%\\
Lateral Inhibition & +0.24\%\\
Graded Inhibition & +0.12\%\\
\hline
Range & 0.12-1.01\%\\
\hline
\end{tabular}
\end{table}

\begin{table}[ht]
\centering
\scriptsize
\caption{Effects of Functional Constraints}
\begin{tabular}{|l|c|}
\hline
Mechanism & Change in Accuracy\\
\hline
Data Augmentation & +22.82\%\\
Convolution layers & +12.14\%\\
LayerNorm & +9.12\%\\
LabelSmoothing & +2.16\%\\
LR Scheduler & +1.01\%\\
\hline
Range & 1.01-22.82\%\\
\hline
\end{tabular}
\end{table}

Through ablation analysis, we found that functional components of the model had a greater effect on model performance than structural components. The most impactful functional constraint(data augmentation, 22.82\%) had 23 times greater of an impact on accuracy than the most impactful structural constraint(connectivity masks, 1.01\%).The structural constraints improved generalization and avoided overfitting, despite having a modest contribution to accuracy. 

\subsection{Limitations}
The limitations of our approach include the approximation of biological principles, constraining the architecture to V1 connectivity, and the lack of representation of visual areas beyond V1. Although the architecture is based on principles from the mouse visual cortex, we did abstract many biological details. The current implementation models only a single cortical column. An accurate reconstruction of the entire mouse visual system would require enhancing the model to include photoreceptors, the lateral geniculate nucleus(LGN), and higher-order visual areas.

Ablation Study Limitations: Our ablation study examines the individual effects of each mechanism used in the BioNIC model, which limits the ability to understand the effects of structural and functional components in tandem. Future works should employ factorial designs which may provide insights into the synergy between structural and functional mechanisms. 

Soft probability distribution used by the regularization mechanism label smoothing aligns more closely with neuroscientific principles than cross entropy loss. However, being a mathematical approach to regularization, it's still not fully biologically truthful. 

In visual input processing, key pre-processing occurs through retinotopic mapping of thalamic input. Here, we implemented convolutional layers in our model, which provide a functionally similar mechanism to retinotopy through operations that preserve spatial relationships in the visual input. In the mouse visual system, higher-order brain regions such as the insular cortex~\cite{salomon-2016}, amygdala~\cite{duncan-2007}, anterior cingulate cortex\cite{choi-2025}, and orbitofrontal cortex~\cite{liu-2020} play critical roles in emotional interpretation. It may be beneficial to extend this approach to include those regions within a more complete model.

\section{Conclusion}
We investigated the potential of incorporating biologically inspired connectivity from the mouse visual cortex into a feedforward neural network for an emotion detection task. We evaluated inter-layer connections, intra-layer connections, graded inhibition, convolutional layer, and regularization. Although the full model did not surpass the performance of the standard model, it offers a promising framework for further exploration of biological-inspired structural constraints in AI models.

\section{Data Availability Statement}

The code and the data supporting the findings in this study are available at:

https://github.com/diyaaprasanth/BioNIC.

\section{Declaration of Generative AI and AI-assisted technologies}
The authors used Perplexity AI and the Gemini-based coding assistant in Google Colab for code refinement and language editing. All outputs were reviewed and revised by the authors, who take full responsibility for the final code and text.

\bibliographystyle{elsarticle-num}
\bibliography{cas-refs}

@book{goodfellow-2013,
	author = {Goodfellow, Ian J. and others},
	booktitle = {Lecture notes in computer science},
	month = {1},
	pages = {117--124},
	title = {{Challenges in Representation Learning: A report on three machine learning contests}},
	year = {2013},
    publisher = {Springer},
	doi = {10.1007/978-3-642-42051-1\_16},
	xurl = {https://doi.org/10.1007/978-3-642-42051-1\_16},
}

@article{bae-2025,
	author = {Bae, J. Alexander and others},
	journal = {Nature},
	month = {4},
	number = {8058},
	pages = {435--447},
	title = {{Functional connectomics spanning multiple areas of mouse visual cortex}},
	volume = {640},
	year = {2025},
	doi = {10.1038/s41586-025-08790-w},
	xurl = {https://doi.org/10.1038/s41586-025-08790-w},
}

@article{dorkenwald-2024,
	author = {Dorkenwald, Sven and others},
	journal = {Nature},
	month = {10},
	number = {8032},
	pages = {124--138},
	title = {{Neuronal wiring diagram of an adult brain}},
	volume = {634},
	year = {2024},
	doi = {10.1038/s41586-024-07558-y},
	xurl = {https://doi.org/10.1038/s41586-024-07558-y},
}

@article{schlegel-2024,
	author = {Schlegel, Philipp and others},
	journal = {Nature},
	month = {10},
	number = {8032},
	pages = {139--152},
	title = {{Whole-brain annotation and multi-connectome cell typing of Drosophila}},
	volume = {634},
	year = {2024},
	doi = {10.1038/s41586-024-07686-5},
	xurl = {https://doi.org/10.1038/s41586-024-07686-5},
}

@article{lappalainen-2024,
	author = {Lappalainen, Janne K and others},
	journal = {Nature},
	month = {9},
	number = {8036},
	pages = {1132--1140},
	title = {{Connectome-constrained networks predict neural activity across the fly visual system}},
	volume = {634},
	year = {2024},
	doi = {10.1038/s41586-024-07939-3},
	xurl = {https://doi.org/10.1038/s41586-024-07939-3},
}

@misc{roberts-2019,
	author = {Roberts, Nicholas and Yap, Dian Ang and Prabhu, Vinay Uday},
	month = {12},
	title = {{Deep Connectomics networks: Neural network architectures inspired by neuronal networks}},
	year = {2019},
	xurl = {https://arxiv.org/abs/1912.08986},
}

@misc{khaireddin-2021,
	author = {Khaireddin, Yousif and Chen, Zhuofa},
	month = {5},
	title = {{Facial Emotion Recognition: state of the art performance on FER2013}},
	year = {2021},
	xurl = {https://arxiv.org/abs/2105.03588},
}

@misc{khanzada-2020,
	author = {Khanzada, Amil and Bai, Charles and Celepcikay, Ferhat Turker},
	month = {4},
	title = {{Facial Expression Recognition with Deep Learning}},
	year = {2020},
	xurl = {https://arxiv.org/abs/2004.11823},
}

@article{tang-2013,
	author = {Tang, Yichuan},
	journal = {arXiv (Cornell University)},
	month = {6},
	title = {{Deep Learning using Support Vector Machines}},
	year = {2013},
	xurl = {https://arxiv.org/pdf/1306.0239},
}

@article{pramerdorfer-2016,
	author = {Pramerdorfer, Christopher and Kampel, Martin},
	journal = {arXiv.org},
	month = {12},
	title = {{Facial Expression Recognition using Convolutional Neural Networks: State of the Art}},
	year = {2016},
	xurl = {https://arxiv.org/abs/1612.02903},
}

@article{goulas-2021,
	author = {Goulas, Alexandros and Damicelli, Fabrizio and Hilgetag, Claus C},
	journal = {bioRxiv (Cold Spring Harbor Laboratory)},
	month = {1},
	title = {{Bio-instantiated recurrent neural networks}},
	year = {2021},
	doi = {10.1101/2021.01.22.427744},
	xurl = {https://doi.org/10.1101/2021.01.22.427744},
}

@article{shi-2022,
	author = {Shi, Jianghong and Tripp, Bryan and Shea-Brown, Eric and Mihalas, Stefan and Buice, Michael A.},
	journal = {PLoS Computational Biology},
	month = {9},
	number = {9},
	pages = {e1010427},
	title = {{MouseNet: A biologically constrained convolutional neural network model for the mouse visual cortex}},
	volume = {18},
	year = {2022},
	doi = {10.1371/journal.pcbi.1010427},
	xurl = {https://doi.org/10.1371/journal.pcbi.1010427},
}

@article{bardozzo-2024,
	author = {Bardozzo, Francesco and Terlizzi, Andrea and Simoncini, Claudio and Lió, Pietro and Tagliaferri, Roberto},
	journal = {Neurocomputing},
	month = {3},
	pages = {127598},
	title = {{Elegans-AI: How the connectome of a living organism could model artificial neural networks}},
	volume = {584},
	year = {2024},
	doi = {10.1016/j.neucom.2024.127598},
	xurl = {https://doi.org/10.1016/j.neucom.2024.127598},
}

@article{damicelli-2022,
	author = {Damicelli, Fabrizio and Hilgetag, Claus C. and Goulas, Alexandros},
	journal = {PLoS Computational Biology},
	month = {11},
	number = {11},
	pages = {e1010639},
	title = {{Brain connectivity meets reservoir computing}},
	volume = {18},
	year = {2022},
	doi = {10.1371/journal.pcbi.1010639},
	xurl = {https://doi.org/10.1371/journal.pcbi.1010639},
}

@article{olivas-2012,
	author = {Olivas, Nicholas D. and Quintanar-Zilinskas, Victor and Nenadic, Zoran and Xu, Xiangmin},
	journal = {Frontiers in Neural Circuits},
	month = {1},
	title = {{Laminar circuit organization and response modulation in mouse visual cortex}},
	volume = {6},
	year = {2012},
	doi = {10.3389/fncir.2012.00070},
	xurl = {https://doi.org/10.3389/fncir.2012.00070},
}

@article{qutub-2023,
	author = {Qutub, Ahmed Adnan Hameed and Atay, Yılmaz},
	journal = {Nexo Revista Científica},
	month = {11},
	number = {05},
	pages = {1--18},
	title = {{Deep Learning Approaches for Classification of Emotion Recognition based on Facial Expressions}},
	volume = {36},
	year = {2023},
	doi = {10.5377/nexo.v36i05.17181},
	xurl = {https://doi.org/10.5377/nexo.v36i05.17181},
}

@article{abdulkadir-2025,
	author = {Abdulkadir, Lujain Y.},
	journal = {Journal of Image and Graphics},
	month = {1},
	number = {5},
	title = {{Evaluating facial emotional proportion based on computer vision technique}},
	volume = {13},
	year = {2025},
	doi = {10.18178/joig.13.5.469-475},
	xurl = {https://doi.org/10.18178/joig.13.5.469-475},
}

@article{dorkenwald-2025,
	author = {Dorkenwald, Sven and others},
	journal = {Nature Methods},
	month = {4},
	number = {5},
	pages = {1112--1120},
	title = {{CAVE: Connectome Annotation Versioning Engine}},
	volume = {22},
	year = {2025},
	doi = {10.1038/s41592-024-02426-z},
	xurl = {https://www.nature.com/articles/s41592-024-02426-z},
}

@article{azevedo-2024,
	author = {Azevedo, Anthony and others},
	journal = {Nature},
	month = {6},
	number = {8020},
	pages = {360--368},
	title = {{Connectomic reconstruction of a female Drosophila ventral nerve cord}},
	volume = {631},
	year = {2024},
	doi = {10.1038/s41586-024-07389-x},
	xurl = {https://doi.org/10.1038/s41586-024-07389-x},
}

@article{elam-2021,
	author = {Elam, Jennifer Stine and others},
	journal = {NeuroImage},
	month = {9},
	pages = {118543},
	title = {{The Human Connectome Project: A retrospective}},
	volume = {244},
	year = {2021},
	doi = {10.1016/j.neuroimage.2021.118543},
	xurl = {https://pubmed.ncbi.nlm.nih.gov/34508893/},
}

@article{salomon-2016,
	author = {Salomon, Roy and others},
	journal = {Journal of Neuroscience},
	month = {5},
	number = {18},
	pages = {5115--5127},
	title = {{The insula mediates access to awareness of visual stimuli presented synchronously to the heartbeat}},
	volume = {36},
	year = {2016},
	doi = {10.1523/jneurosci.4262-15.2016},
	xurl = {https://doi.org/10.1523/jneurosci.4262-15.2016},
}

@article{duncan-2007,
	author = {Duncan, Seth and Barrett, Lisa Feldman},
	journal = {Trends in Cognitive Sciences},
	month = {4},
	number = {5},
	pages = {190--192},
	title = {{The role of the amygdala in visual awareness}},
	volume = {11},
	year = {2007},
	doi = {10.1016/j.tics.2007.01.007},
	xurl = {https://doi.org/10.1016/j.tics.2007.01.007},
}

@article{choi-2025,
	author = {Choi, Jiye and Lee, Young-Beom and So, Dahm and Kim, Jee Yeon and Choi, Sungjoon and Kim, Sowon and Keum, Sehoon},
	journal = {Nature Communications},
	month = {2},
	number = {1},
	pages = {1937},
	title = {{Cortical representations of affective pain shape empathic fear in male mice}},
	volume = {16},
	year = {2025},
	doi = {10.1038/s41467-025-57230-w},
	xurl = {https://www.nature.com/articles/s41467-025-57230-w},
}

@article{liu-2020,
	author = {Liu, Dechen and Deng, Juan and Zhang, Zhewei and Zhang, Zhi-Yu and Sun, Yan-Gang and Yang, Tianming and Yao, Haishan},
	journal = {Nature Communications},
	month = {6},
	number = {1},
	pages = {2784},
	title = {{Orbitofrontal control of visual cortex gain promotes visual associative learning}},
	volume = {11},
	year = {2020},
	doi = {10.1038/s41467-020-16609-7},
	xurl = {https://doi.org/10.1038/s41467-020-16609-7},
}

@article{ding-2025,
	author = {Ding, Zhuokun and others},
	journal = {Nature},
	month = {4},
	number = {8058},
	pages = {459--469},
	title = {{Functional connectomics reveals general wiring rule in mouse visual cortex}},
	volume = {640},
	year = {2025},
	doi = {10.1038/s41586-025-08840-3},
	xurl = {https://doi.org/10.1038/s41586-025-08840-3},
}

@book{busse-2018,
	author = {Busse, Laura},
	booktitle = {Handbook of behavioral neuroscience},
	month = {1},
    publisher = {Elsevier},
	pages = {53--68},
	title = {{The mouse visual system and visual perception}},
	year = {2018},
	doi = {10.1016/b978-0-12-812012-5.00004-5},
	xurl = {https://doi.org/10.1016/b978-0-12-812012-5.00004-5},
}

@misc{caveclient-author-no-date,
	title = {{CAVEclient 1.0 documentation}},
	url = {https://caveclient.readthedocs.io},
    year = {2025},
}

@article{bodor-2025,
	author = {Bodor, Agnes L and others},
	journal = {Nature Neuroscience},
	month = {7},
	number = {8},
	pages = {1704--1715},
	title = {{The synaptic architecture of layer 5 thick tufted excitatory neurons in mouse visual cortex}},
	volume = {28},
	year = {2025},
	doi = {10.1038/s41593-025-02004-2},
	xurl = {https://doi.org/10.1038/s41593-025-02004-2},
}

@article{simonyan-2022,
	author = {Simonyan, Karen and Zisserman, Andrew},
	journal = {arXiv (Cornell University)},
	month = {3},
	title = {{Very deep convolutional networks for Large-Scale image recognition}},
	year = {2022},
	doi = {10.48550/arxiv.1409.1556},
	xurl = {http://arxiv.org/abs/1409.1556},
}

@article{mollahosseini-2022,
	author = {Mollahosseini, Ali and Chan, David and Mahoor, Mohammad H.},
	journal = {arXiv (Cornell University)},
	month = {3},
	title = {{Going Deeper in Facial Expression Recognition using Deep Neural Networks}},
	year = {2022},
	doi = {10.48550/arxiv.1511.04110},
	xurl = {https://doi.org/10.48550/arxiv.1511.04110},
}

@misc{ Ngai-2025,
	month = {4},
	title = {{From the BRAIN Director: How a cubic millimeter of brain can Change Neuroscience | BRAIN Initiative}},
	year = {2025},
	xurl = {https://braininitiative.nih.gov/news-events/blog/brain-director-how-cubic-millimeter-brain-can-change-neuroscience},
}

@article{kingma-2014,
	author = {Kingma, Diederik P. and Ba, Jimmy},
	journal = {arXiv (Cornell University)},
	month = {12},
	title = {{Adam: A method for stochastic optimization}},
	year = {2014},
	doi = {10.48550/arxiv.1412.6980},
	xurl = {http://arxiv.org/abs/1412.6980},
}

@article{shorten-2019,
	author = {Shorten, Connor and Khoshgoftaar, Taghi M.},
	journal = {Journal Of Big Data},
	month = {7},
	number = {1},
	title = {{A survey on Image Data Augmentation for Deep Learning}},
	volume = {6},
	year = {2019},
	doi = {10.1186/s40537-019-0197-0},
	xurl = {https://doi.org/10.1186/s40537-019-0197-0},
}

@article{mao-2024,
	author = {Mao, Xiaoyi and Staiger, Jochen F.},
	journal = {Pflügers Archiv - European Journal of Physiology},
	month = {2},
	number = {5},
	pages = {721--733},
	title = {{Multimodal cortical neuronal cell type classification}},
	volume = {476},
	year = {2024},
	doi = {10.1007/s00424-024-02923-2},
	xurl = {https://doi.org/10.1007/s00424-024-02923-2},
}

@misc{ReduceLROnPlateau-2023,
	author = {Contributors, PyTorch},
	month = {1},
	title = {{ReduceLROnPlateau}},
	year = {2023},
	xurl = {https://pytorch.org/docs/stable/generated/torch.optim.lr_scheduler.ReduceLROnPlateau.html},
}

@article{suto-2024,
	author = {Suto, Jozsef},
	journal = {Sensors},
	month = {7},
	number = {14},
	pages = {4502},
	title = {{Using data augmentation to improve the generalization capability of an object detector on Remote-Sensed Insect Trap images}},
	volume = {24},
	year = {2024},
	doi = {10.3390/s24144502},
	xurl = {https://pmc.ncbi.nlm.nih.gov/articles/PMC11281147/},
}

@article{szegedy-2016,
	author = {Szegedy, Christian and Google Inc. and Vanhoucke, Vincent and Google Inc. and Ioffe, Sergey and Google Inc. and Shlens, Jon and Google Inc. and Wojna, Zbigniew and University College London},
	journal = {IEEE Xplore},
	title = {{Rethinking the Inception Architecture for Computer Vision}},
	year = {2016},
	xurl = {https://www.cv-foundation.org/openaccess/content_cvpr_2016/papers/Szegedy_Rethinking_the_Inception_CVPR_2016_paper.pdf},
}

@techreport{muller-no-date,
	author = {Müller, Rafael and Kornblith, Simon and Hinton, Geoffrey and Google Brain},
	title = {{When does label smoothing help?}},
	xurl = {https://proceedings.neurips.cc/paper_files/paper/2019/file/f1748d6b0fd9d439f71450117eba2725-Paper.pdf},
    year = {2019},
    institution = {Google Brain},
}

@techreport{hensch-2004,
	author = {Hensch, Takao K. and Laboratory for Neuronal Circuit Development, Critical Period Mechanisms Research Group, RIKEN Brain Science Institute},
	pages = {549--579},
	title = {{CRITICAL PERIOD REGULATION}},
	year = {2004},
	doi = {10.1146/annurev.neuro.27.070203.144327},
	xurl = {https://henschlab.mcb.harvard.edu/wp-content/uploads/2012/06/hensch-ann-rev-neurosci-2004.pdf},
    institution = {Harvard University}
}

@article{wiesel-1963,
	author = {Wiesel, Torsten N. and Hubel, David H.},
	journal = {Journal of Neurophysiology},
	month = {11},
	number = {6},
	pages = {1003--1017},
	title = {{SINGLE-CELL RESPONSES IN STRIATE CORTEX OF KITTENS DEPRIVED OF VISION IN ONE EYE}},
	volume = {26},
	year = {1963},
	doi = {10.1152/jn.1963.26.6.1003},
	xurl = {https://doi.org/10.1152/jn.1963.26.6.1003},
}

@article{raviv-2022,
	author = {Raviv, Limor and Lupyan, Gary and Green, Shawn C.},
	journal = {Trends in Cognitive Sciences},
	month = {5},
	number = {6},
	pages = {462--483},
	title = {{How variability shapes learning and generalization}},
	volume = {26},
	year = {2022},
	doi = {10.1016/j.tics.2022.03.007},
	xurl = {https://doi.org/10.1016/j.tics.2022.03.007},
}

@article{hernandez-garcia-2020,
	author = {Hernández-García, Alex},
	journal = {arXiv (Cornell University)},
	month = {12},
	title = {{Data augmentation and image understanding}},
	year = {2020},
	doi = {10.48550/arxiv.2012.14185},
	xurl = {http://arxiv.org/abs/2012.14185},
}

@article{shen-2021,
	author = {Shen, Yang and Wang, Julia and Navlakha, Saket},
	journal = {Neural Computation},
	month = {9},
	number = {12},
	pages = {3179--3203},
	title = {{A correspondence between normalization strategies in artificial and biological neural networks}},
	volume = {33},
	year = {2021},
	doi = {10.1162/neco\_a\_01439},
	xurl = {https://doi.org/10.1162/neco\_a\_01439},
}

@article{lian-2023,
	author = {Lian, Yanbo and Burkitt, Anthony N.},
	journal = {bioRxiv (Cold Spring Harbor Laboratory)},
	month = {6},
	title = {{Relating sparse/predictive coding to divisive normalization}},
	year = {2023},
	doi = {10.1101/2023.06.08.544285},
	xurl = {https://doi.org/10.1101/2023.06.08.544285},
}

@article{hu-2018,
	author = {Hu, Jie and Shen, Li and Albanie, Samuel and Sun, Gang and Wu, Enhua},
	journal = {IEEE},
	month = {6},
	pages = {7132--7141},
	title = {{Squeeze-and-Excitation Networks}},
	year = {2018},
	doi = {10.1109/cvpr.2018.00745},
	xurl = {https://doi.org/10.1109/cvpr.2018.00745},
}

@book{woo-2018,
	author = {Woo, Sanghyun and Park, Jongchan and Lee, Joon-Young and Kweon, In So},
	booktitle = {Lecture notes in computer science},
	month = {1},
	pages = {3--19},
	title = {{CBAM: Convolutional Block Attention Module}},
    publisher = {arXiv},
	year = {2018},
	doi = {10.1007/978-3-030-01234-2\_1},
	xurl = {https://doi.org/10.1007/978-3-030-01234-2_1},
}

@article{he-2016,
	author = {He, Kaiming and Zhang, Xiangyu and Ren, Shaoqing and Sun, Jian},
	journal = {IEEE/CVF Conference on Computer Vision and Pattern Recognition (CVPR)},
	month = {6},
	pages = {770--778},
	title = {{Deep Residual Learning for Image Recognition}},
	year = {2016},
	doi = {10.1109/cvpr.2016.90},
	xurl = {https://doi.org/10.1109/cvpr.2016.90},
}

@article{nair-2010,
	author = {Nair, Vinod and Hinton, Geoffrey E.},
	journal = {International Conference on Machine Learning},
	month = {6},
	pages = {807--814},
	title = {{Rectified linear units improve restricted Boltzmann machines}},
	year = {2010},
	xurl = {https://icml.cc/Conferences/2010/papers/432.pdf},
}

@book{prechelt-1998,
	author = {Prechelt, Lutz},
	booktitle = {Lecture notes in computer science},
	month = {1},
	pages = {55--69},
	title = {{Early stopping - but when?}},
    publisher = {Springer},
	year = {1998},
	doi = {10.1007/3-540-49430-8\_3},
	xurl = {https://doi.org/10.1007/3-540-49430-8_3},
}

@article{krogh-1991,
	author = {Krogh, Anders and Hertz, John A.},
	journal = {NIPS},
	month = {12},
	pages = {950--957},
	title = {{A Simple Weight Decay Can Improve Generalization}},
	volume = {4},
	year = {1991},
	xurl = {http://citeseerx.ist.psu.edu/viewdoc/summary?doi=10.1.1.41.2305},
}

@article{fawcett-2005,
	author = {Fawcett, Tom},
	journal = {Pattern Recognition Letters},
	month = {12},
	number = {8},
	pages = {861--874},
	title = {{An introduction to ROC analysis}},
	volume = {27},
	year = {2005},
	doi = {10.1016/j.patrec.2005.10.010},
	xurl = {https://doi.org/10.1016/j.patrec.2005.10.010},
}

@misc{wandell-1995,
	author = {Wandell, Brian},
	title = {{Foundations of Vision (1995)}},
    year = {1995},
	url = {https://wandell.github.io/FOV-1995/},
}

@article{ioffe-2015,
	author = {Ioffe, Sergey and Szegedy, Christian},
	journal = {arXiv (Cornell University)},
	month = {2},
	title = {{Batch normalization: Accelerating deep network training by reducing internal covariate shift}},
	year = {2015},
	doi = {10.48550/arxiv.1502.03167},
	xurl = {http://arxiv.org/abs/1502.03167},
}

@article{krizhevsky-2017,
	author = {Krizhevsky, Alex and Sutskever, Ilya and Hinton, Geoffrey E.},
	journal = {Communications of the ACM},
	month = {5},
	number = {6},
	pages = {84--90},
	title = {{ImageNet classification with deep convolutional neural networks}},
	volume = {60},
	year = {2017},
	doi = {10.1145/3065386},
	xurl = {https://doi.org/10.1145/3065386},
}

@article{schneider-mizell-2025,
	author = {Schneider-Mizell, Casey M and others},
	journal = {Nature},
	month = {4},
	number = {8058},
	pages = {448--458},
	title = {{Inhibitory specificity from a connectomic census of mouse visual cortex}},
	volume = {640},
	year = {2025},
	doi = {10.1038/s41586-024-07780-8},
	xurl = {https://www.nature.com/articles/s41586-024-07780-8},
}

@article{morris-1999,
	author = {Morris, R.G.M},
	journal = {Brain Research Bulletin},
	month = {11},
	number = {5-6},
	pages = {437},
	title = {{D.O. Hebb: The Organization of Behavior, Wiley: New York; 1949}},
	volume = {50},
	year = {1999},
	doi = {10.1016/s0361-9230(99)00182-3},
	xurl = {https://doi.org/10.1016/s0361-9230(99)00182-3},
}

@article{turrigiano-2000,
	author = {Turrigiano, Gina G and Nelson, Sacha B},
	journal = {Current Opinion in Neurobiology},
	month = {6},
	number = {3},
	pages = {358--364},
	title = {{Hebb and homeostasis in neuronal plasticity}},
	volume = {10},
	year = {2000},
	doi = {10.1016/s0959-4388(00)00091-x},
	xurl = {https://doi.org/10.1016/s0959-4388(00)00091-x},
}

@article{zenke-2013,
	author = {Zenke, Friedemann and Hennequin, Guillaume and Gerstner, Wulfram},
	journal = {PLoS Computational Biology},
	month = {11},
	number = {11},
	pages = {e1003330},
	title = {{Synaptic Plasticity in Neural Networks Needs Homeostasis with a Fast Rate Detector}},
	volume = {9},
	year = {2013},
	doi = {10.1371/journal.pcbi.1003330},
	xurl = {https://doi.org/10.1371/journal.pcbi.1003330},
}

\clearpage
\appendix
\begin{figure*}[t]
\centering
{\Large\bfseries Appendix A. Confusion matrix, and ROC Curves for the models used in the ablation study - Last run of each ablation experiment}\par
\vspace{1em}

\noindent\begin{minipage}[t]{0.44\textwidth}
\includegraphics[height=0.23\textheight,width=\textwidth,keepaspectratio]{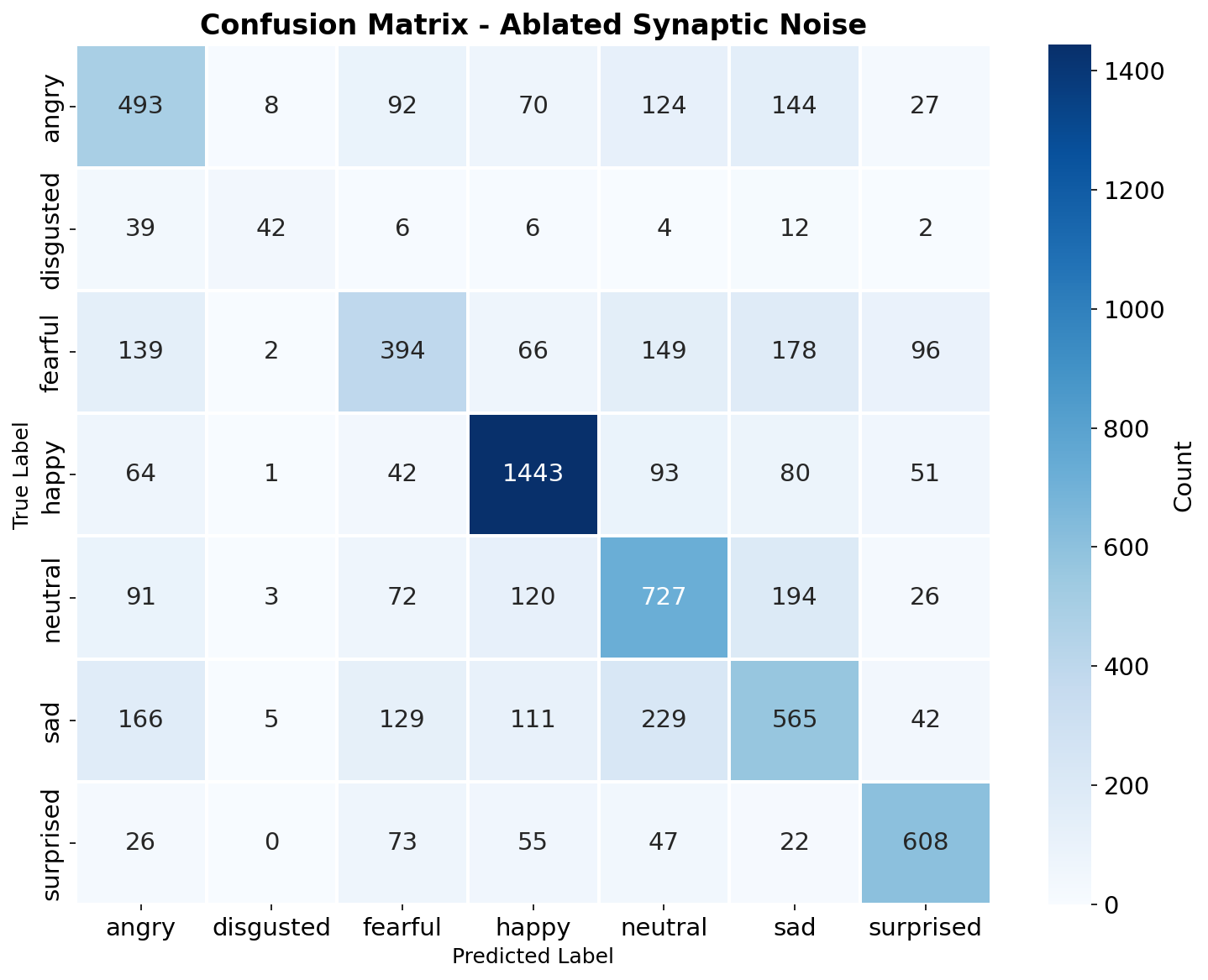}
\end{minipage}%
\hfill
\begin{minipage}[t]{0.44\textwidth}
\includegraphics[height=0.23\textheight,width=\textwidth,keepaspectratio]{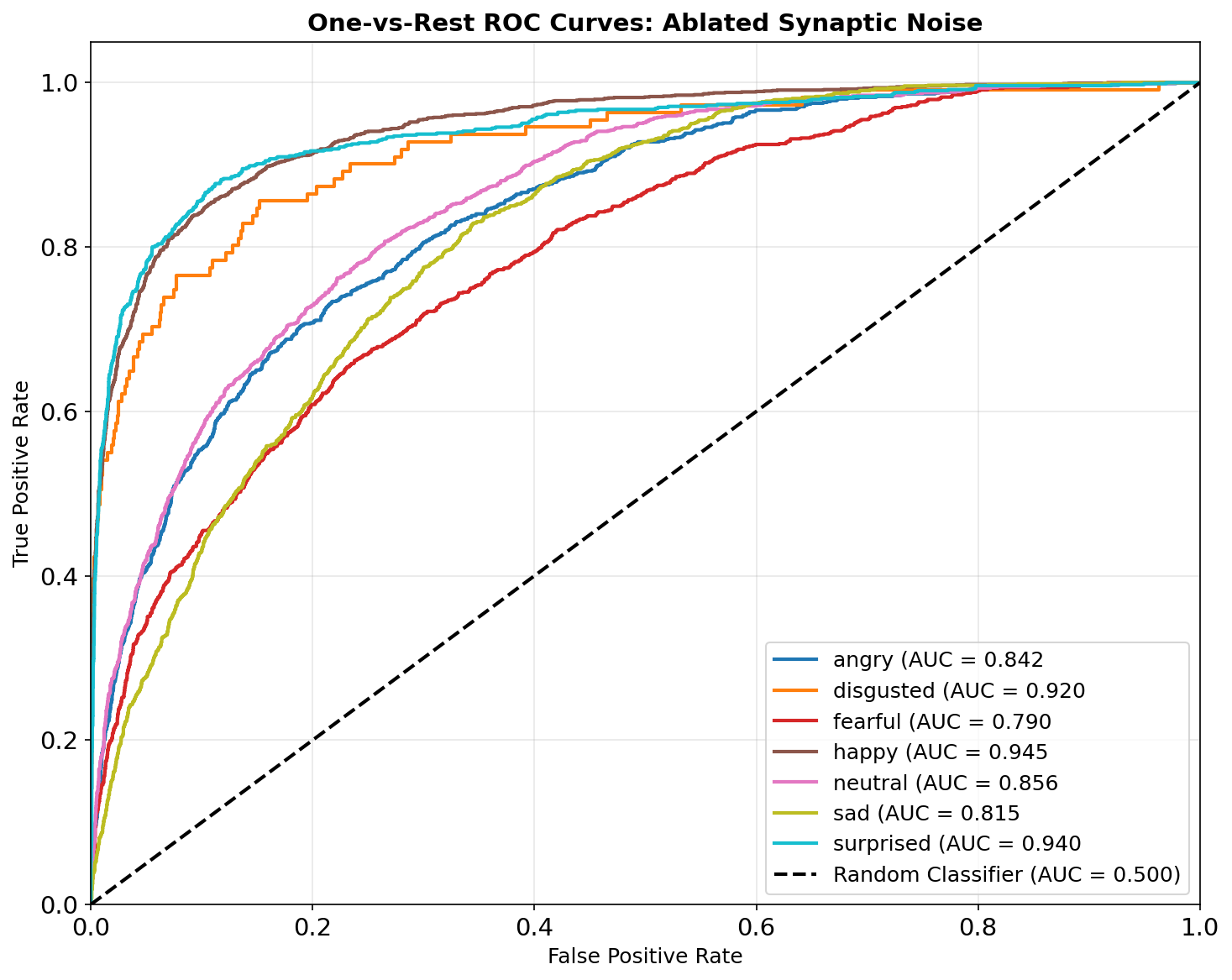}
\end{minipage}

\vspace{0.2em}

\noindent\begin{minipage}[t]{0.44\textwidth}
\includegraphics[height=0.23\textheight,width=\textwidth,keepaspectratio]{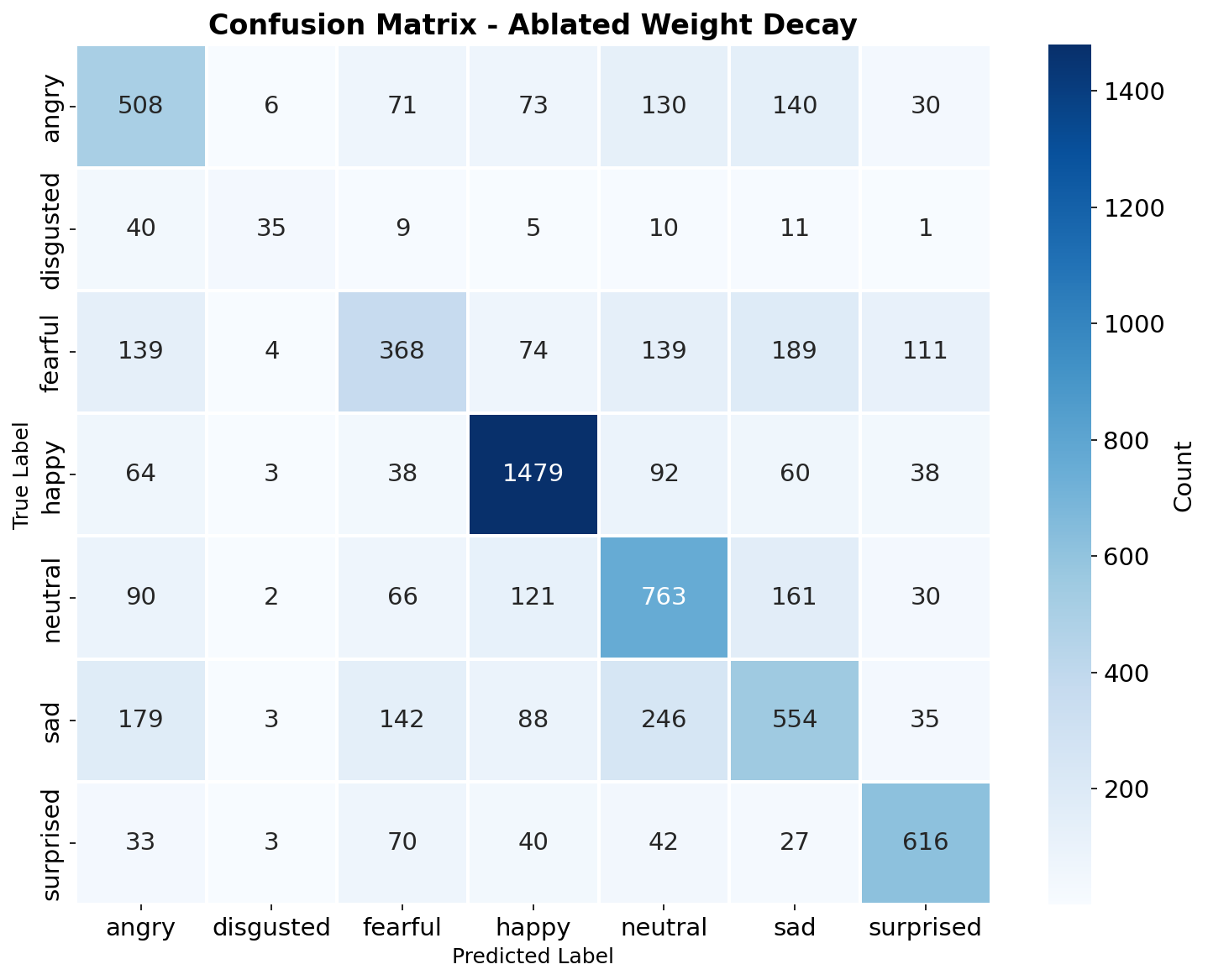}
\end{minipage}%
\hfill
\begin{minipage}[t]{0.44\textwidth}
\includegraphics[height=0.23\textheight,width=\textwidth,keepaspectratio]{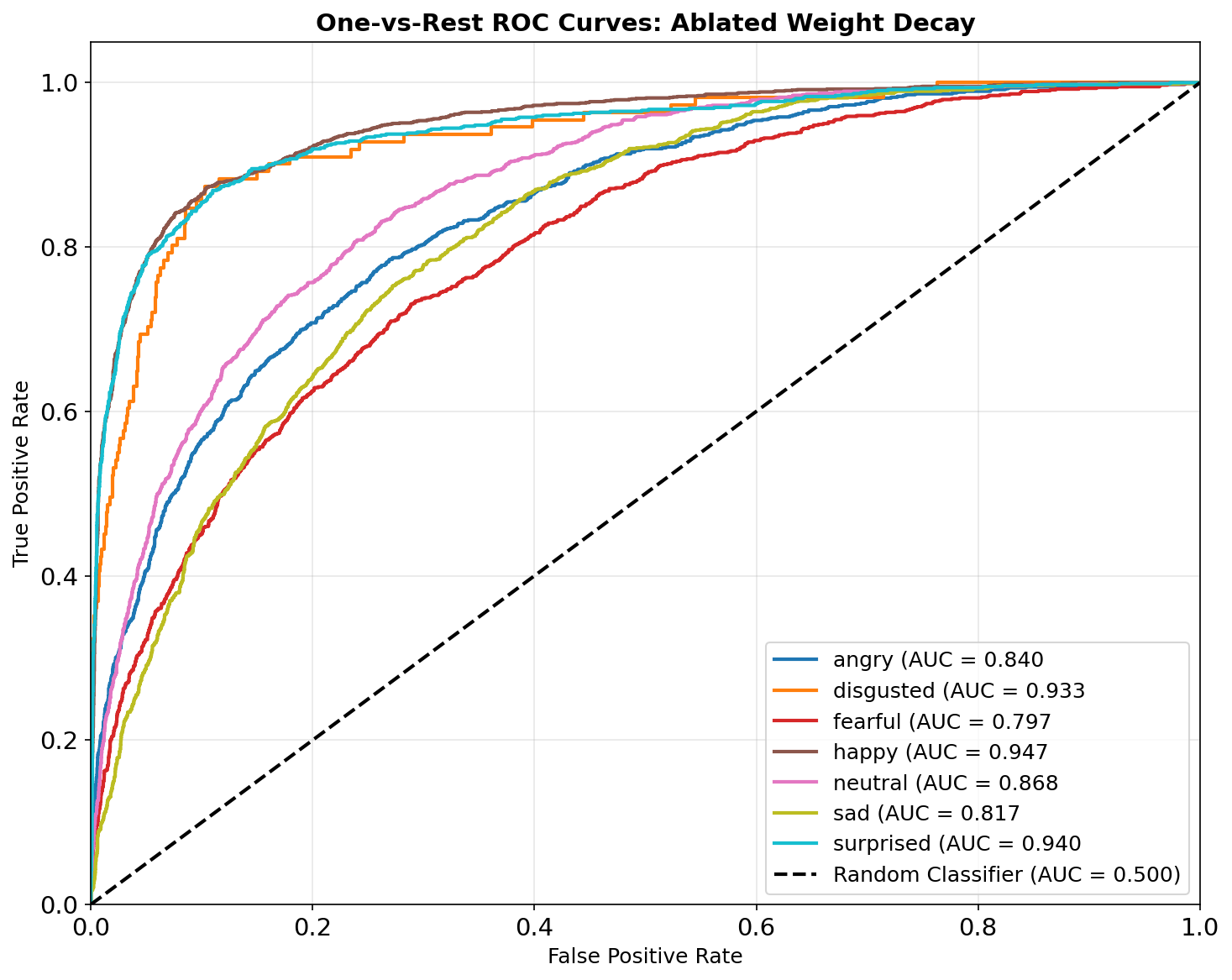}
\end{minipage}

\vspace{0.2em}

\noindent\begin{minipage}[t]{0.44\textwidth}
\includegraphics[height=0.23\textheight,width=\textwidth,keepaspectratio]{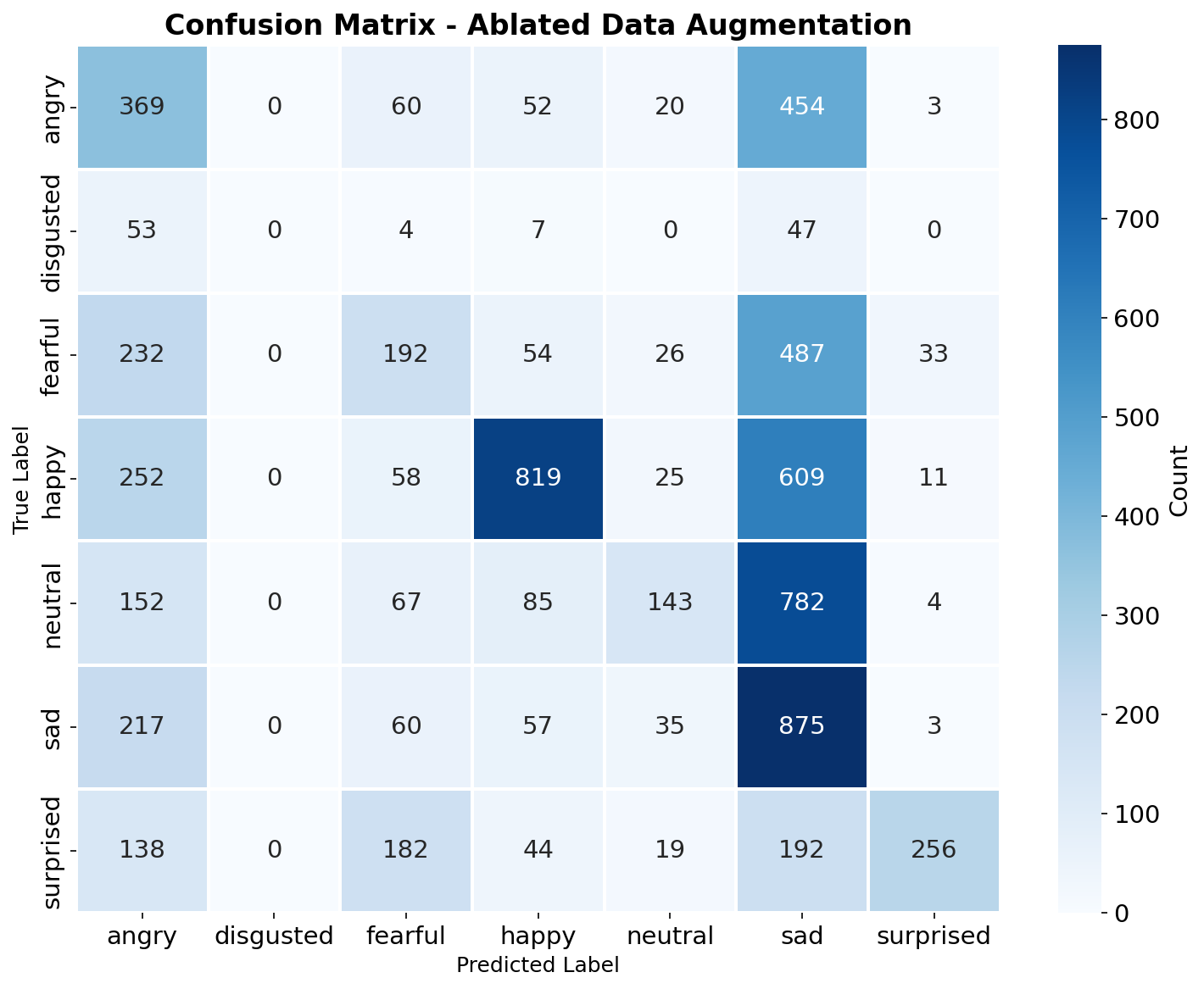}
\end{minipage}%
\hfill
\begin{minipage}[t]{0.44\textwidth}
\includegraphics[height=0.23\textheight,width=\textwidth,keepaspectratio]{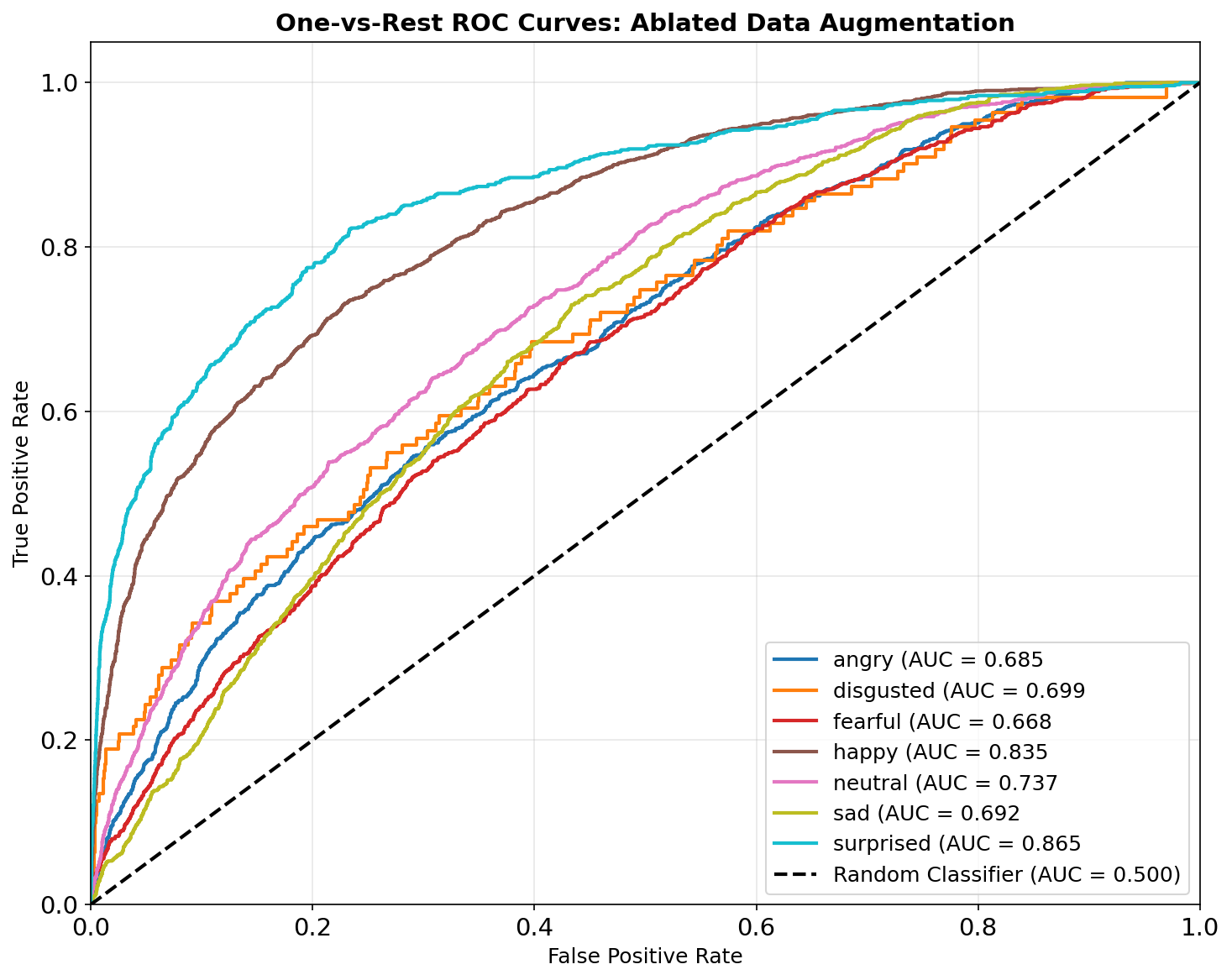}
\end{minipage}

\vspace{0.2em}

\noindent\begin{minipage}[t]{0.44\textwidth}
\includegraphics[height=0.23\textheight,width=\textwidth,keepaspectratio]{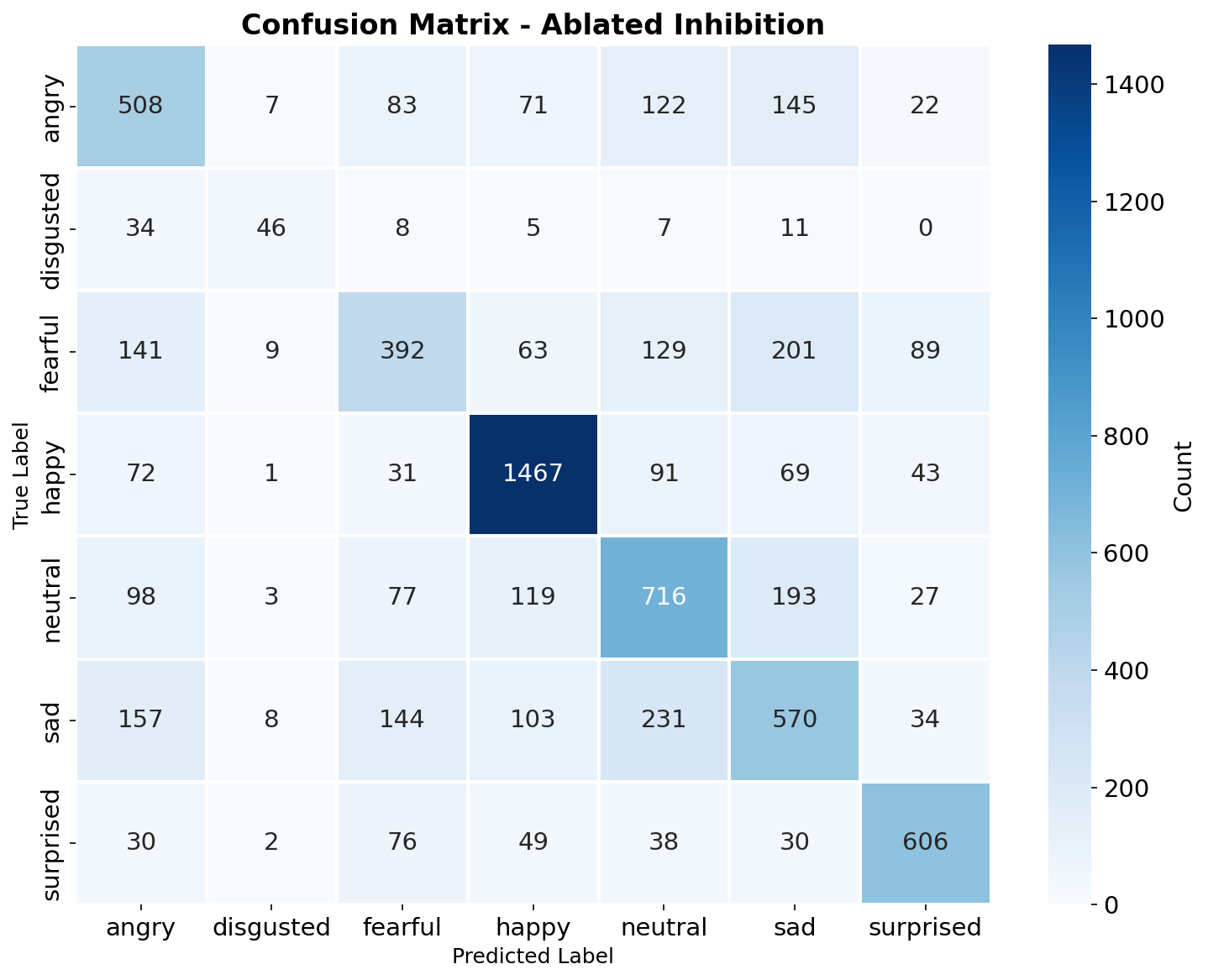}
\end{minipage}%
\hfill
\begin{minipage}[t]{0.44\textwidth}
\includegraphics[height=0.23\textheight,width=\textwidth,keepaspectratio]{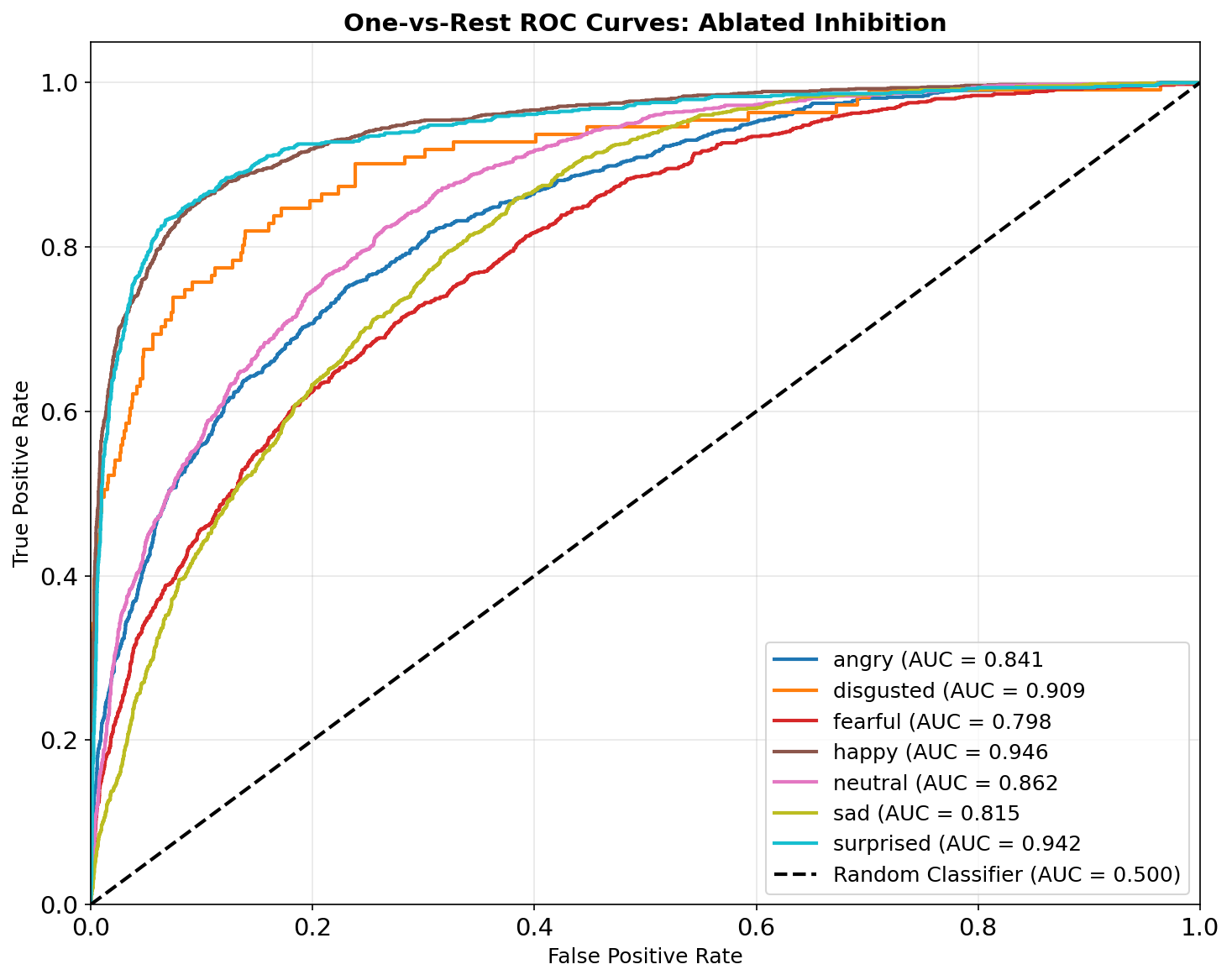}
\end{minipage}

\end{figure*}

\begin{figure*}[t]

\noindent\begin{minipage}[t]{0.44\textwidth}
\includegraphics[height=0.23\textheight,width=\textwidth,keepaspectratio]{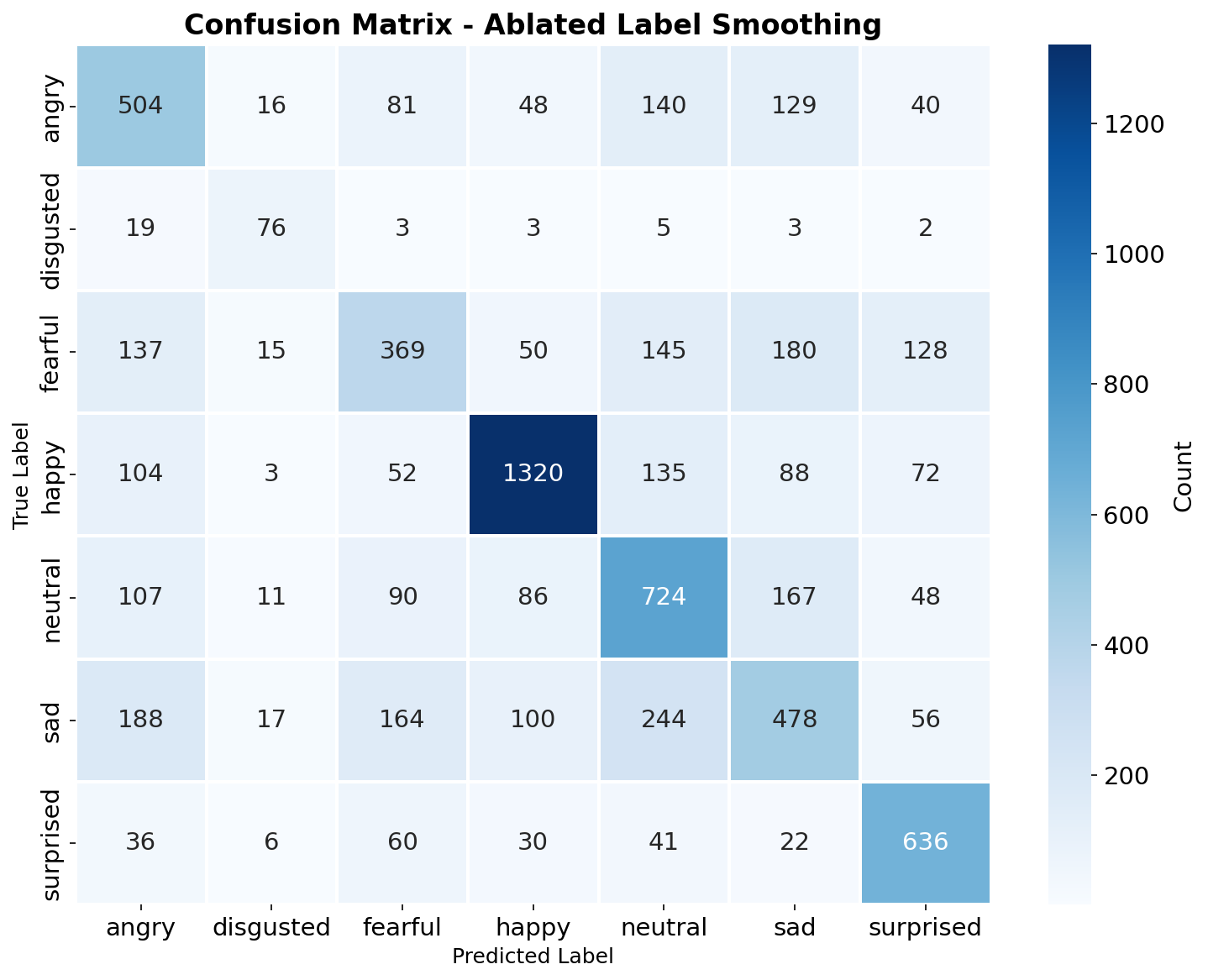}
\end{minipage}%
\hfill
\begin{minipage}[t]{0.44\textwidth}
\includegraphics[height=0.23\textheight,width=\textwidth,keepaspectratio]{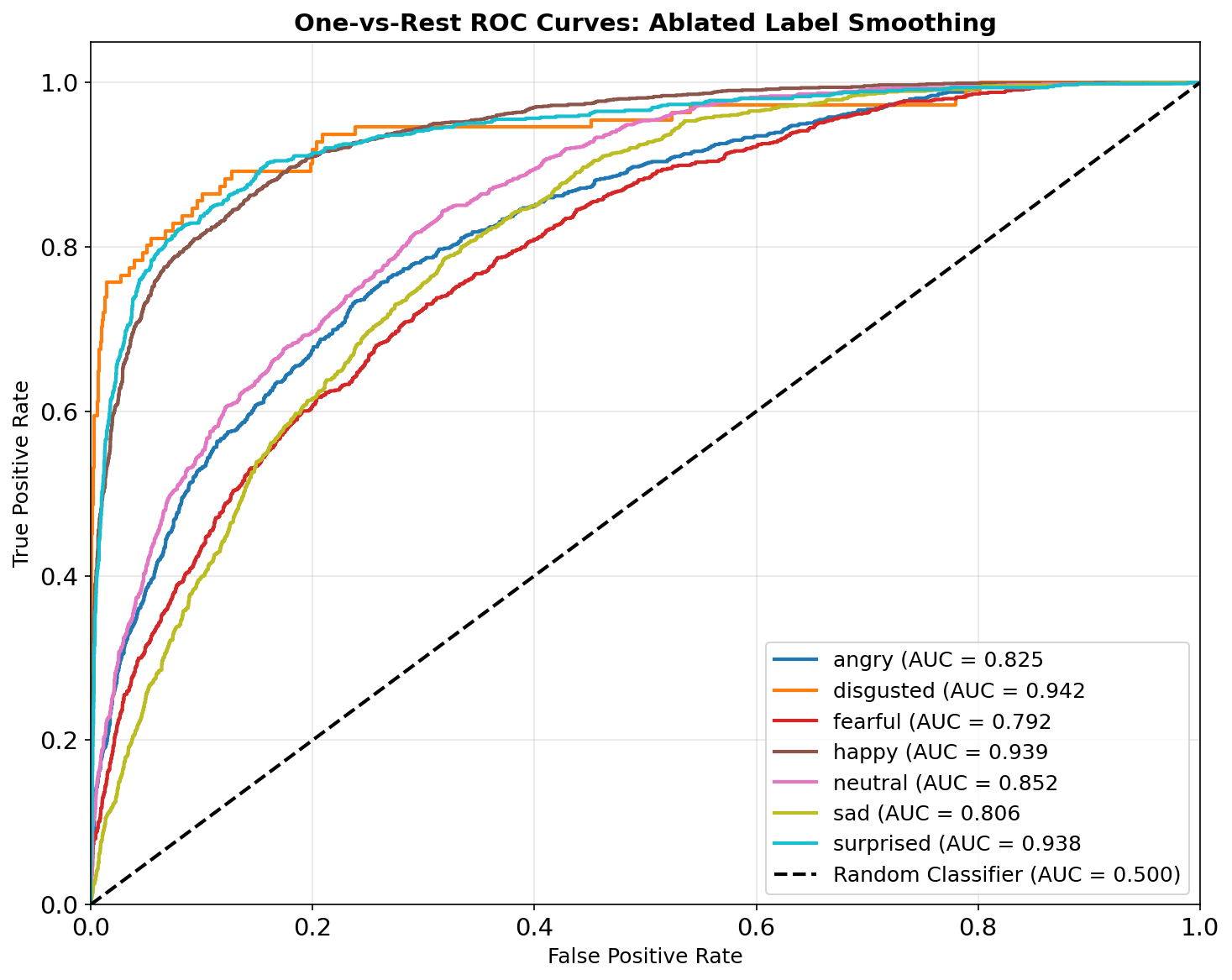}
\end{minipage}

\vspace{0.2em}

\noindent\begin{minipage}[t]{0.44\textwidth}
\includegraphics[height=0.23\textheight,width=\textwidth,keepaspectratio]{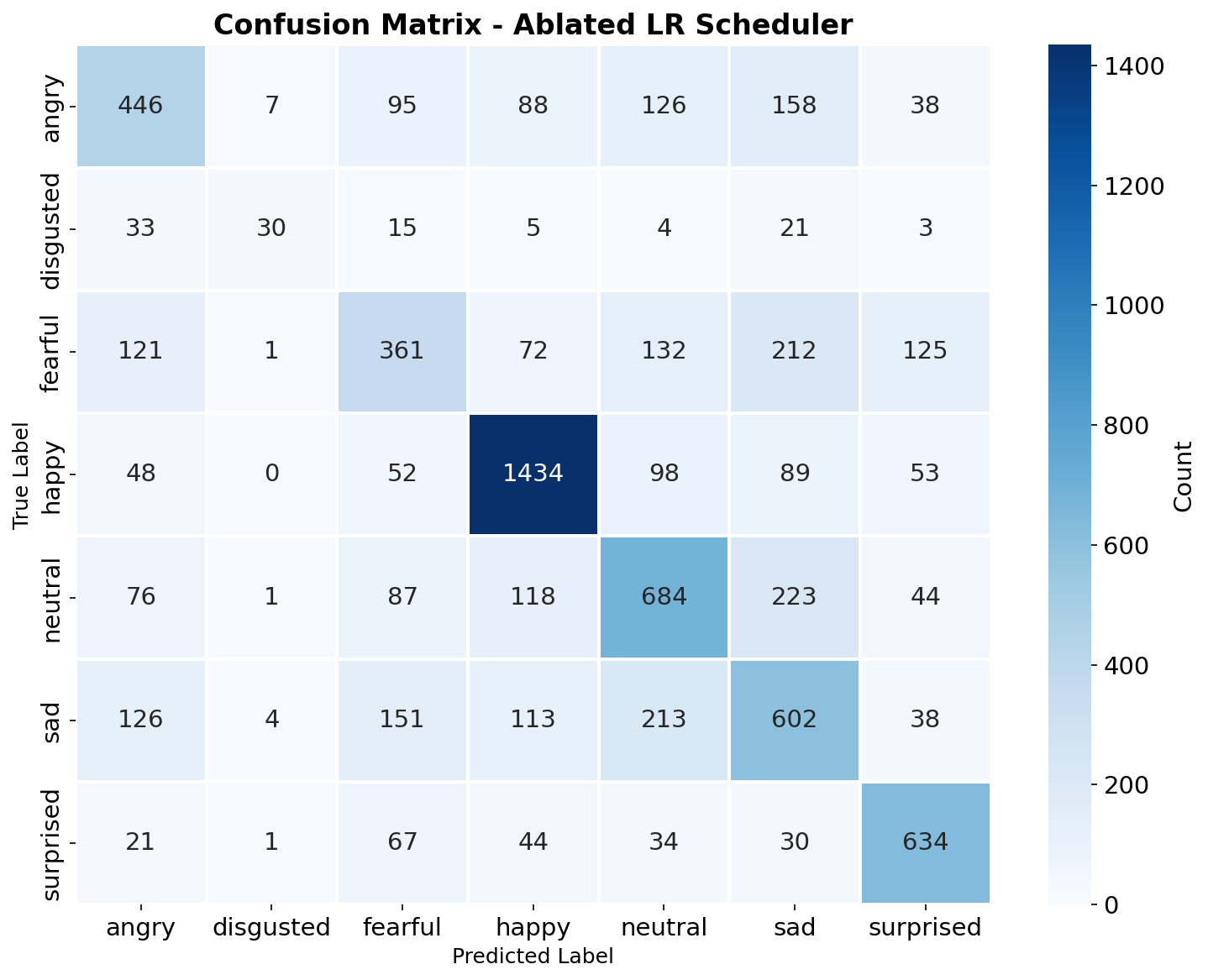}
\end{minipage}%
\hfill
\begin{minipage}[t]{0.44\textwidth}
\includegraphics[height=0.23\textheight,width=\textwidth,keepaspectratio]{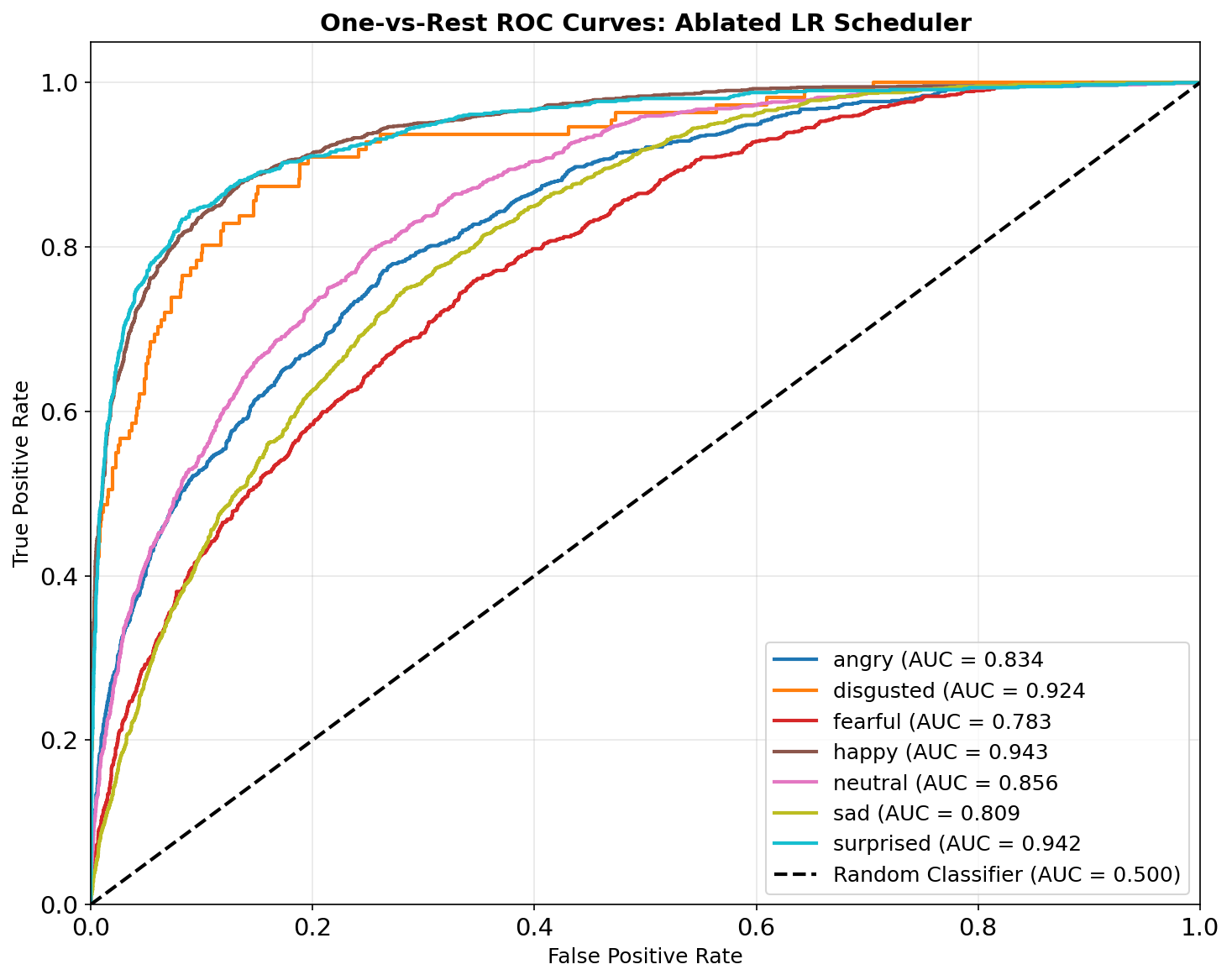}
\end{minipage}

\vspace{0.2em}

\noindent\begin{minipage}[t]{0.44\textwidth}
\includegraphics[height=0.23\textheight,width=\textwidth,keepaspectratio]{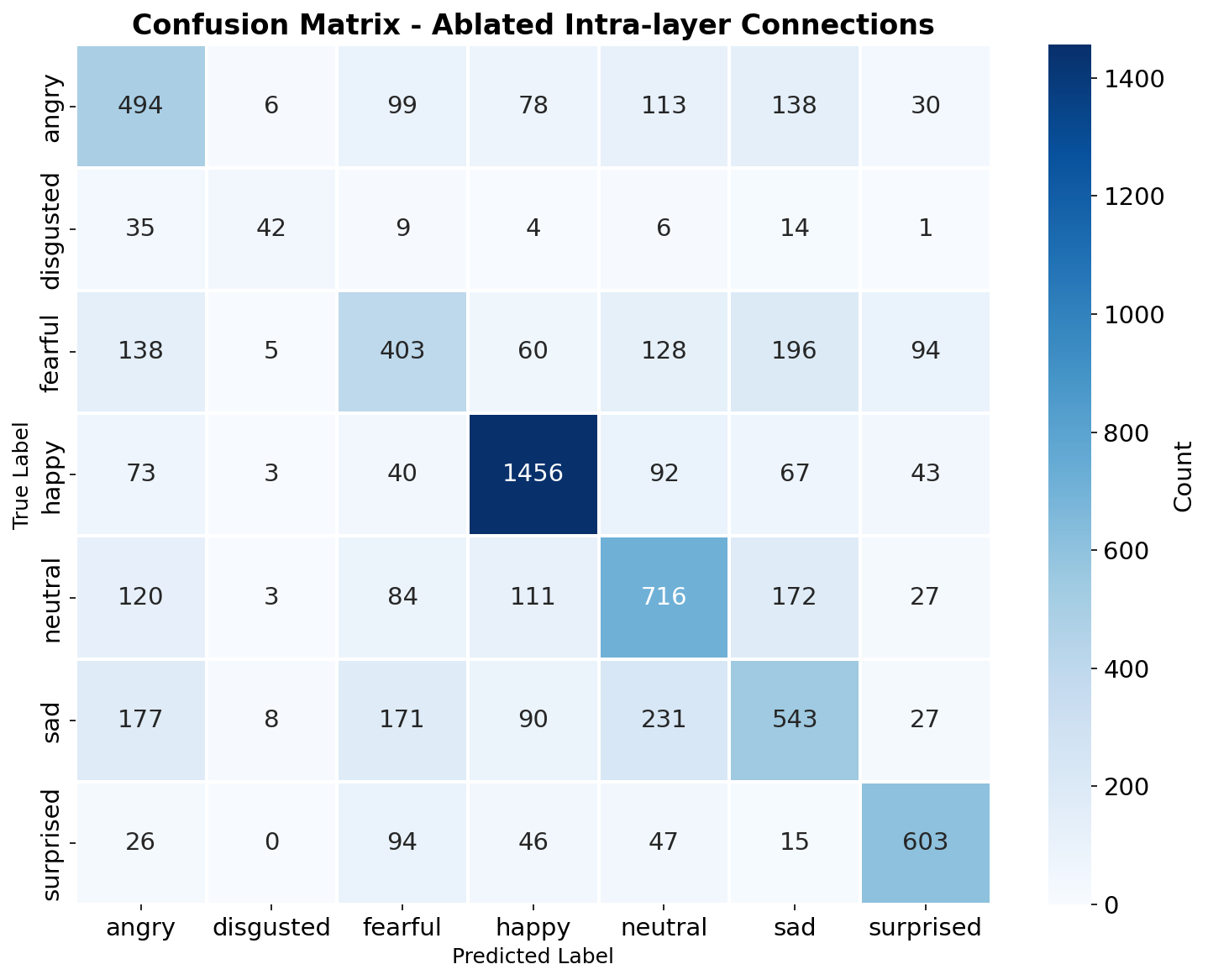}
\end{minipage}%
\hfill
\begin{minipage}[t]{0.44\textwidth}
\includegraphics[height=0.23\textheight,width=\textwidth,keepaspectratio]{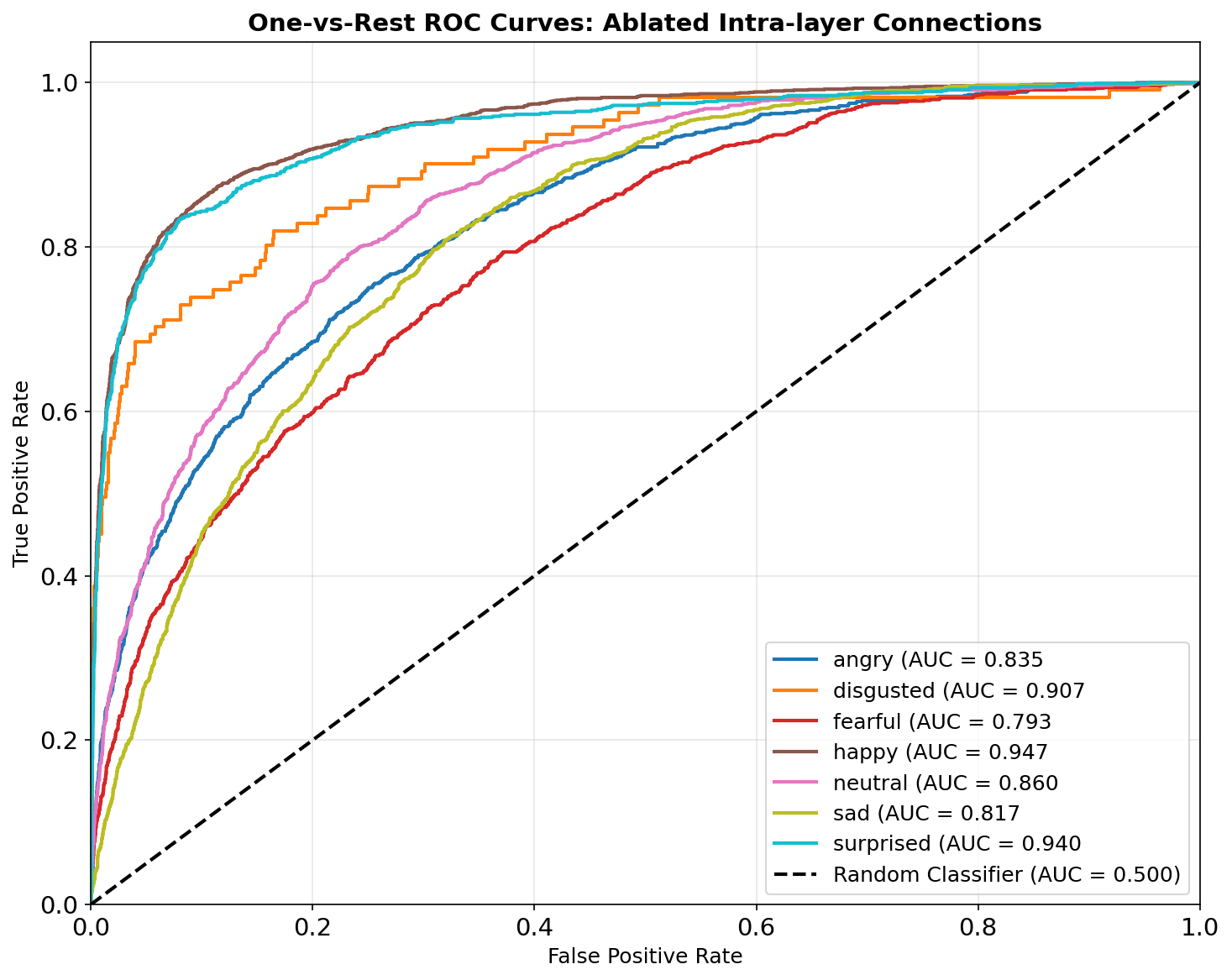}
\end{minipage}

\vspace{0.2em}

\noindent\begin{minipage}[t]{0.44\textwidth}
\includegraphics[height=0.23\textheight,width=\textwidth,keepaspectratio]{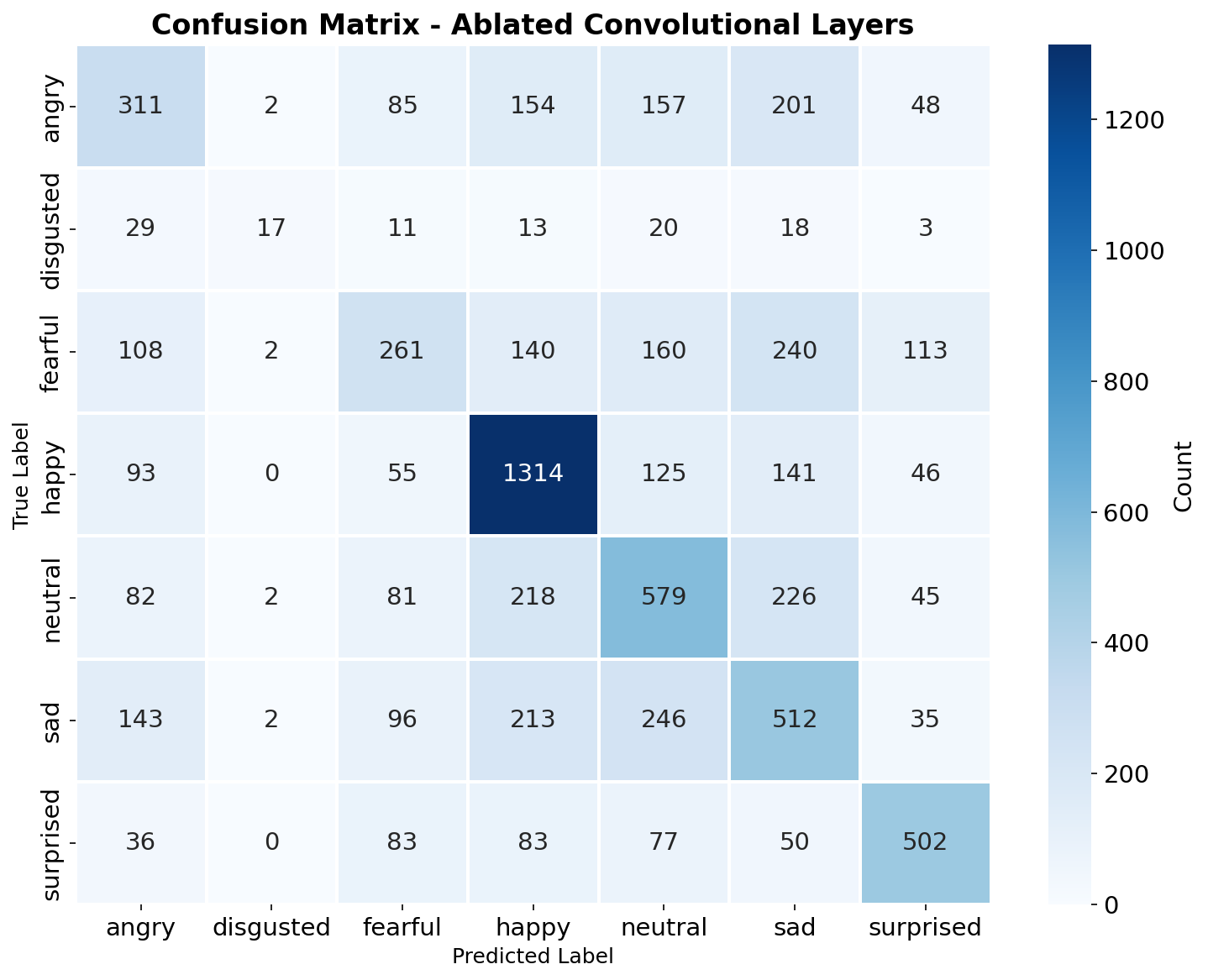}
\end{minipage}%
\hfill
\begin{minipage}[t]{0.44\textwidth}
\includegraphics[height=0.23\textheight,width=\textwidth,keepaspectratio]{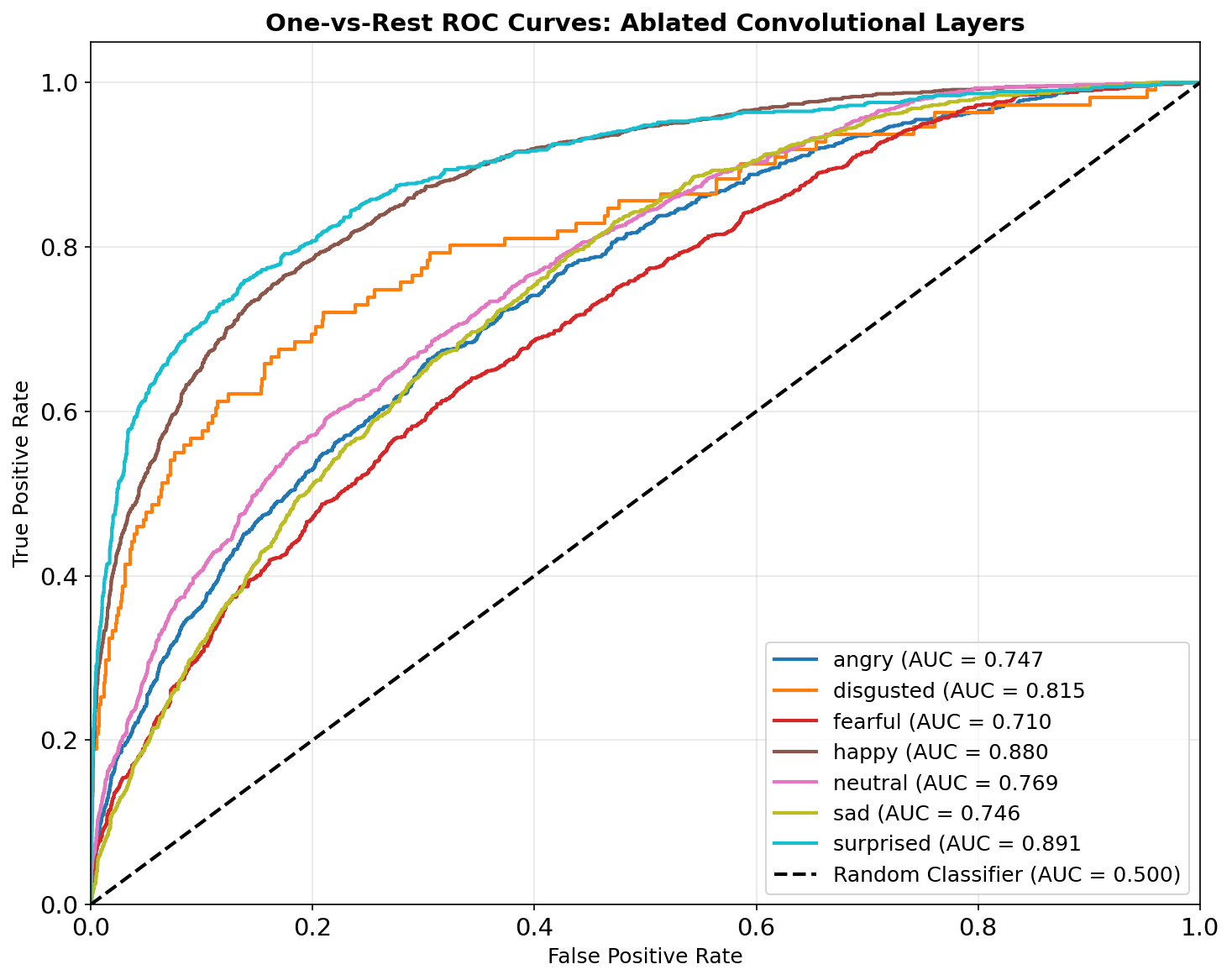}
\end{minipage}

\end{figure*}

\begin{figure*}[t]

\noindent\begin{minipage}[t]{0.44\textwidth}
\includegraphics[height=0.24\textheight,width=\textwidth,keepaspectratio]{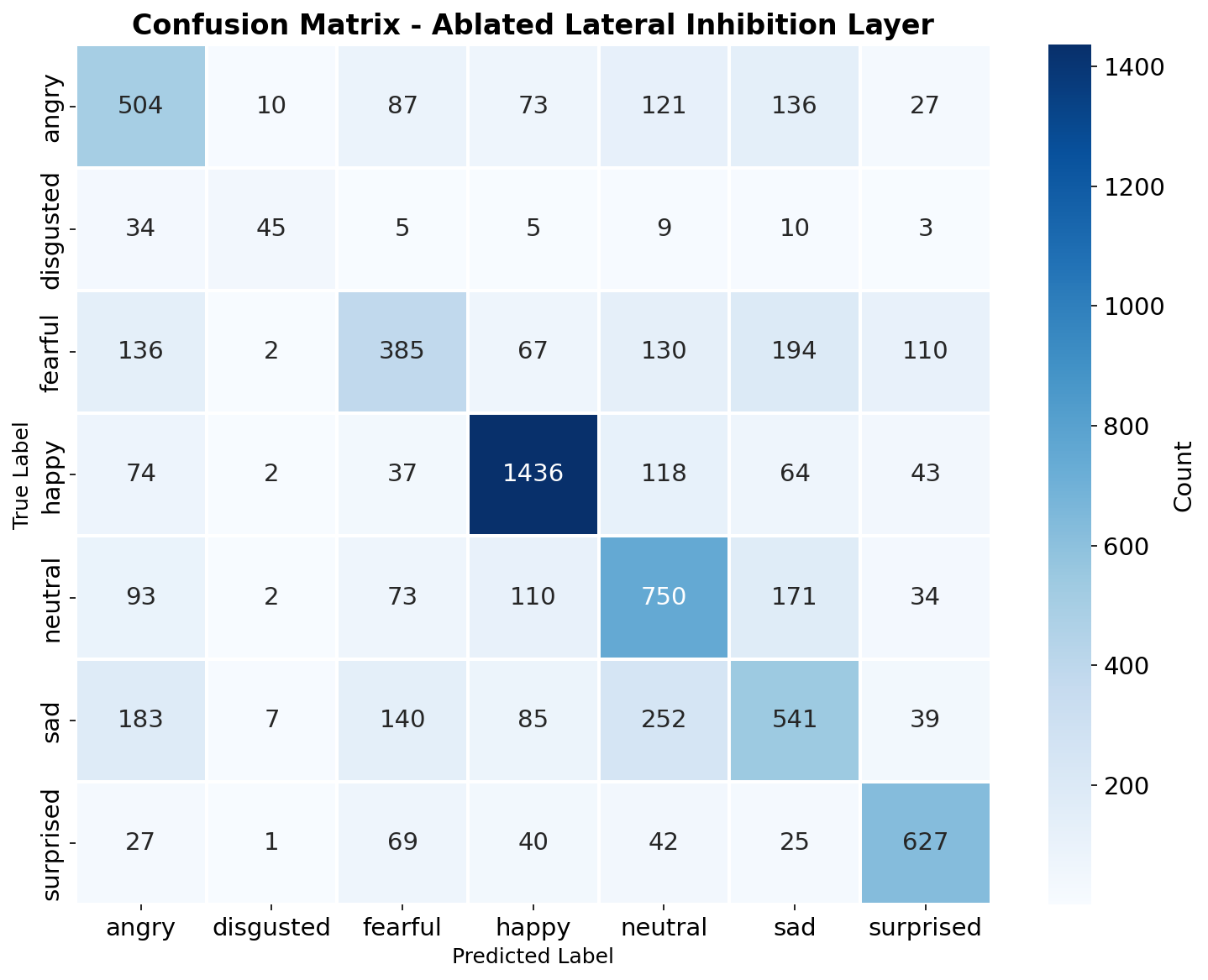}
\end{minipage}%
\hfill
\begin{minipage}[t]{0.44\textwidth}
\includegraphics[height=0.24\textheight,width=\textwidth,keepaspectratio]{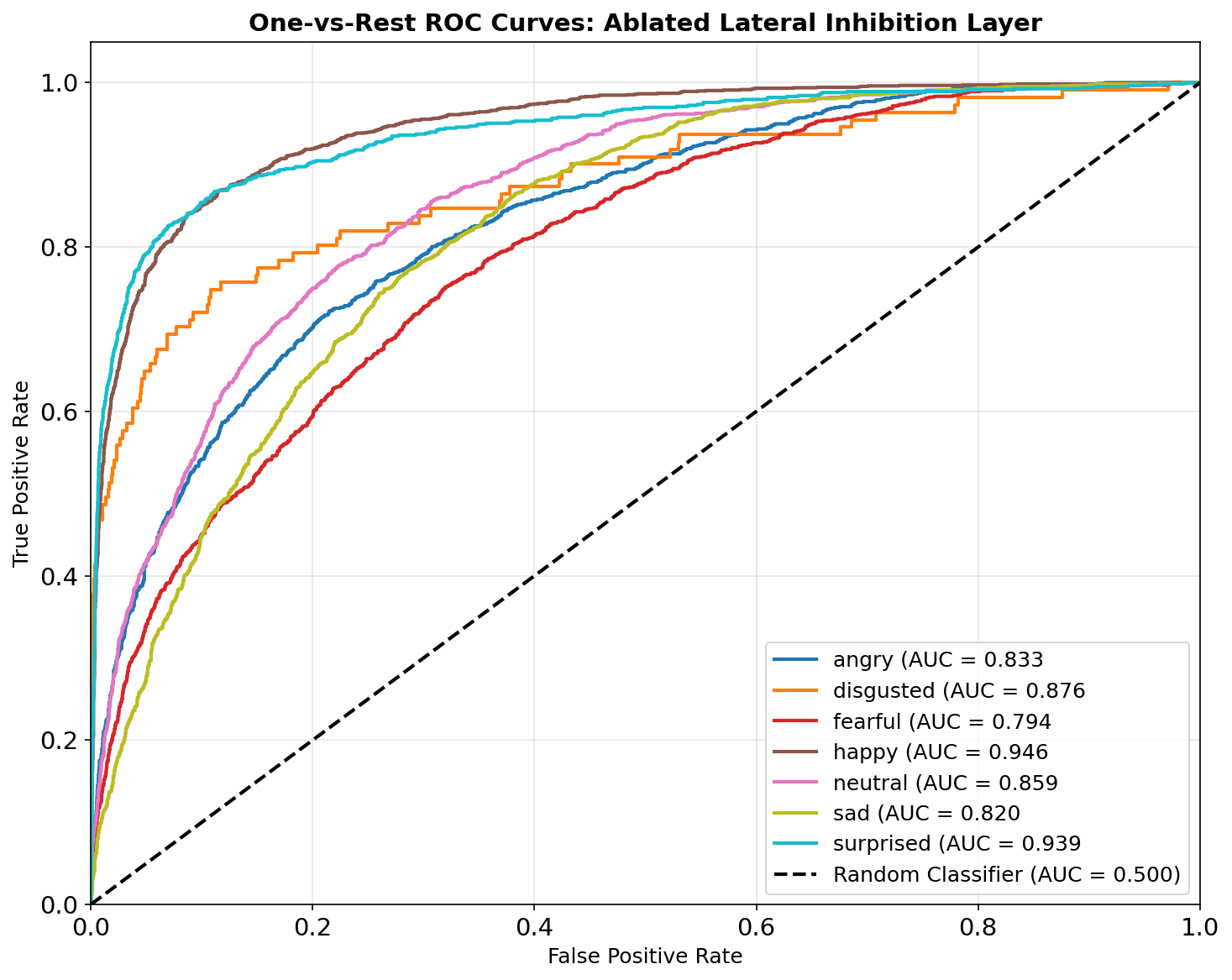}
\end{minipage}

\vspace{0.3em}

\noindent\begin{minipage}[t]{0.44\textwidth}
\includegraphics[height=0.24\textheight,width=\textwidth,keepaspectratio]{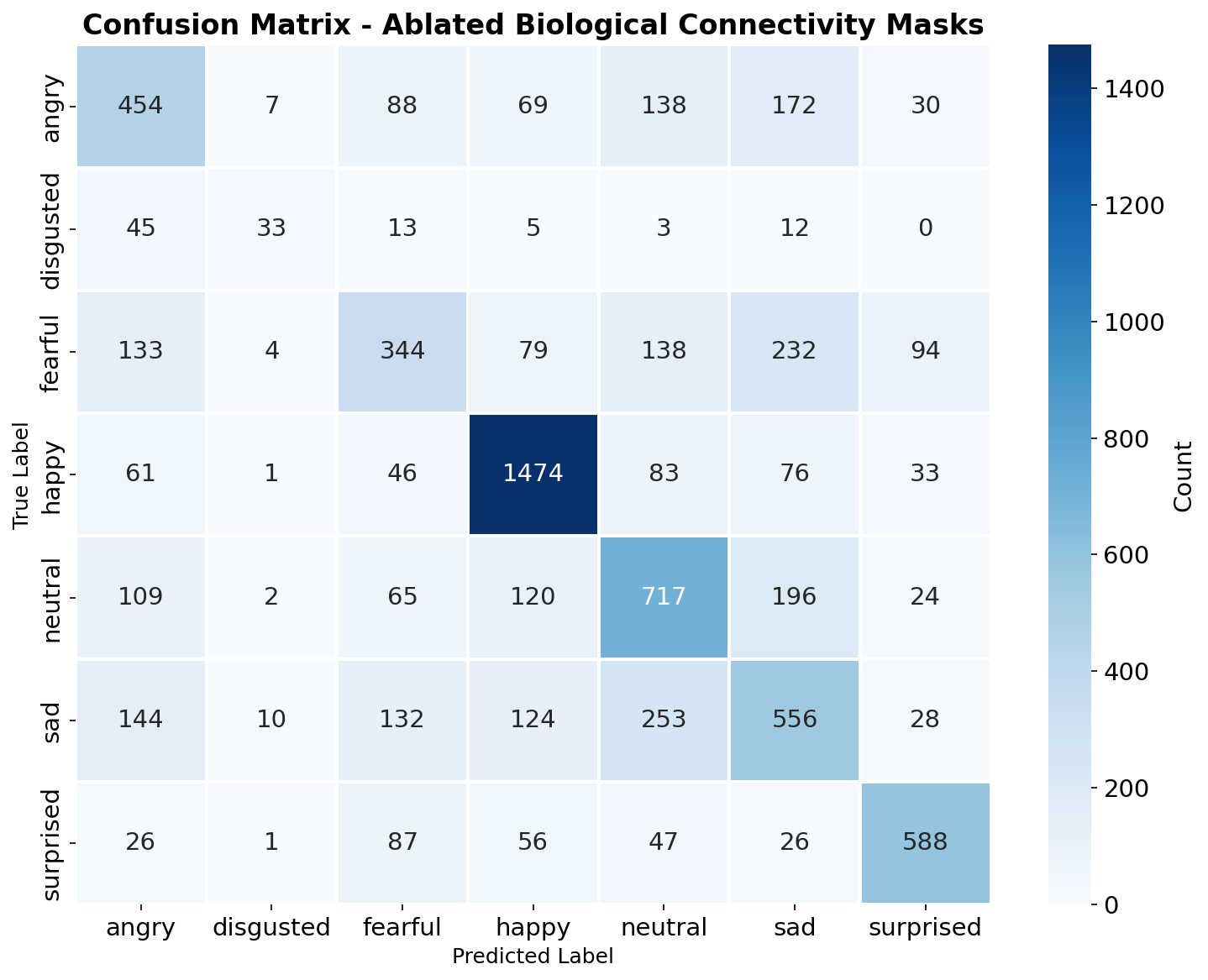}
\end{minipage}%
\hfill
\begin{minipage}[t]{0.44\textwidth}
\includegraphics[height=0.24\textheight,width=\textwidth,keepaspectratio]{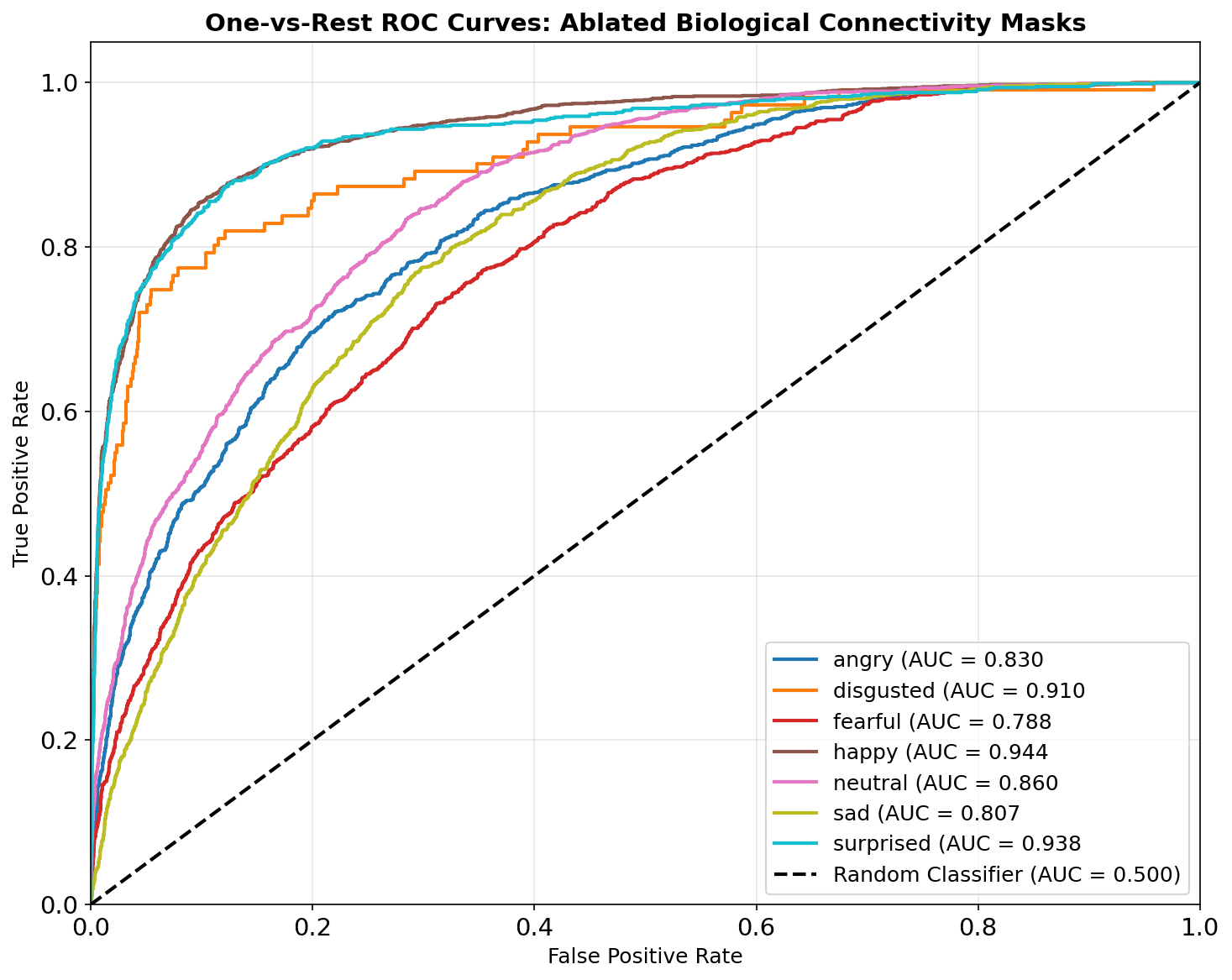}
\end{minipage}

\vspace{0.3em}

\noindent\begin{minipage}[t]{0.44\textwidth}
\includegraphics[height=0.24\textheight,width=\textwidth,keepaspectratio]{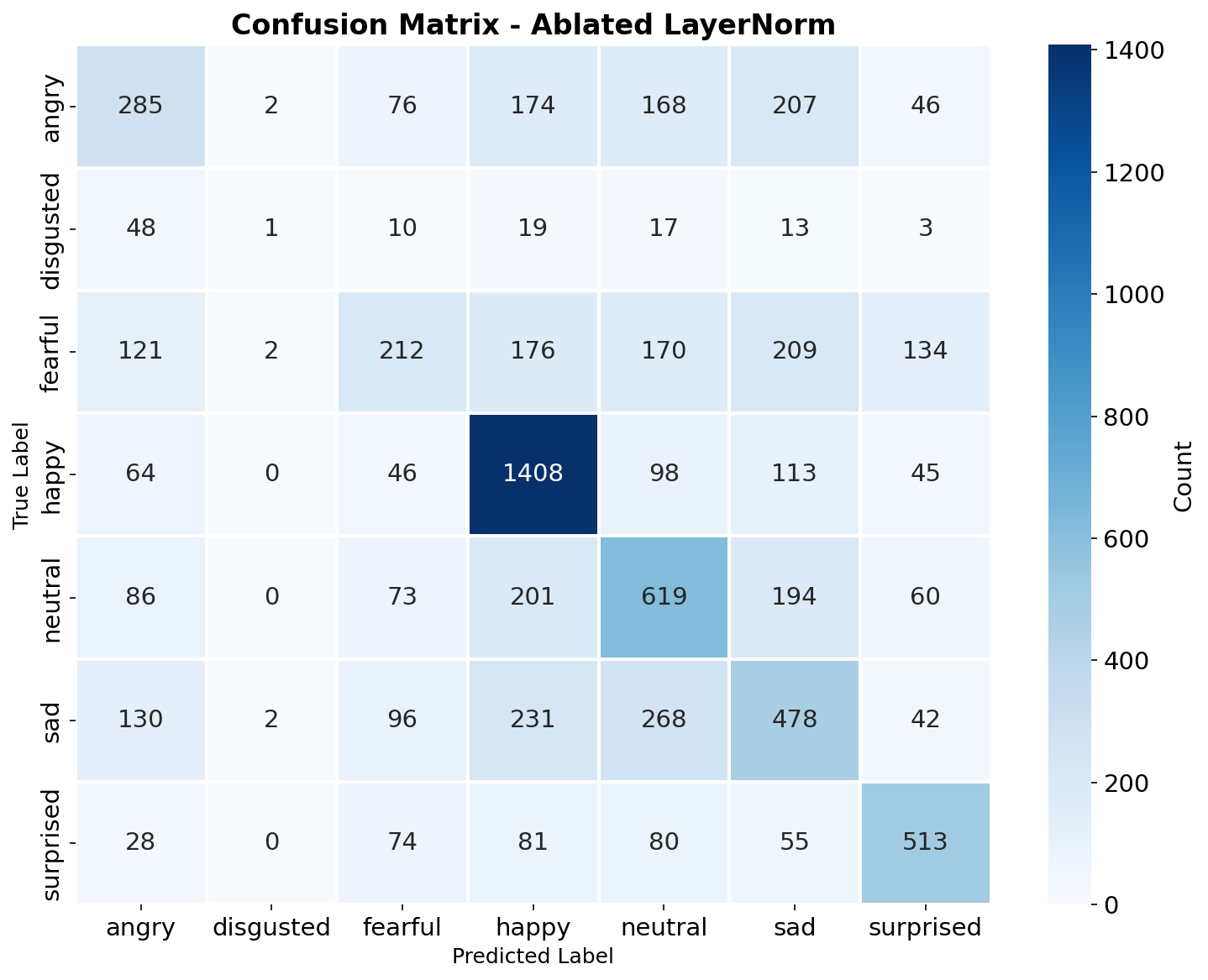}
\end{minipage}%
\hfill
\begin{minipage}[t]{0.44\textwidth}
\includegraphics[height=0.24\textheight,width=\textwidth,keepaspectratio]{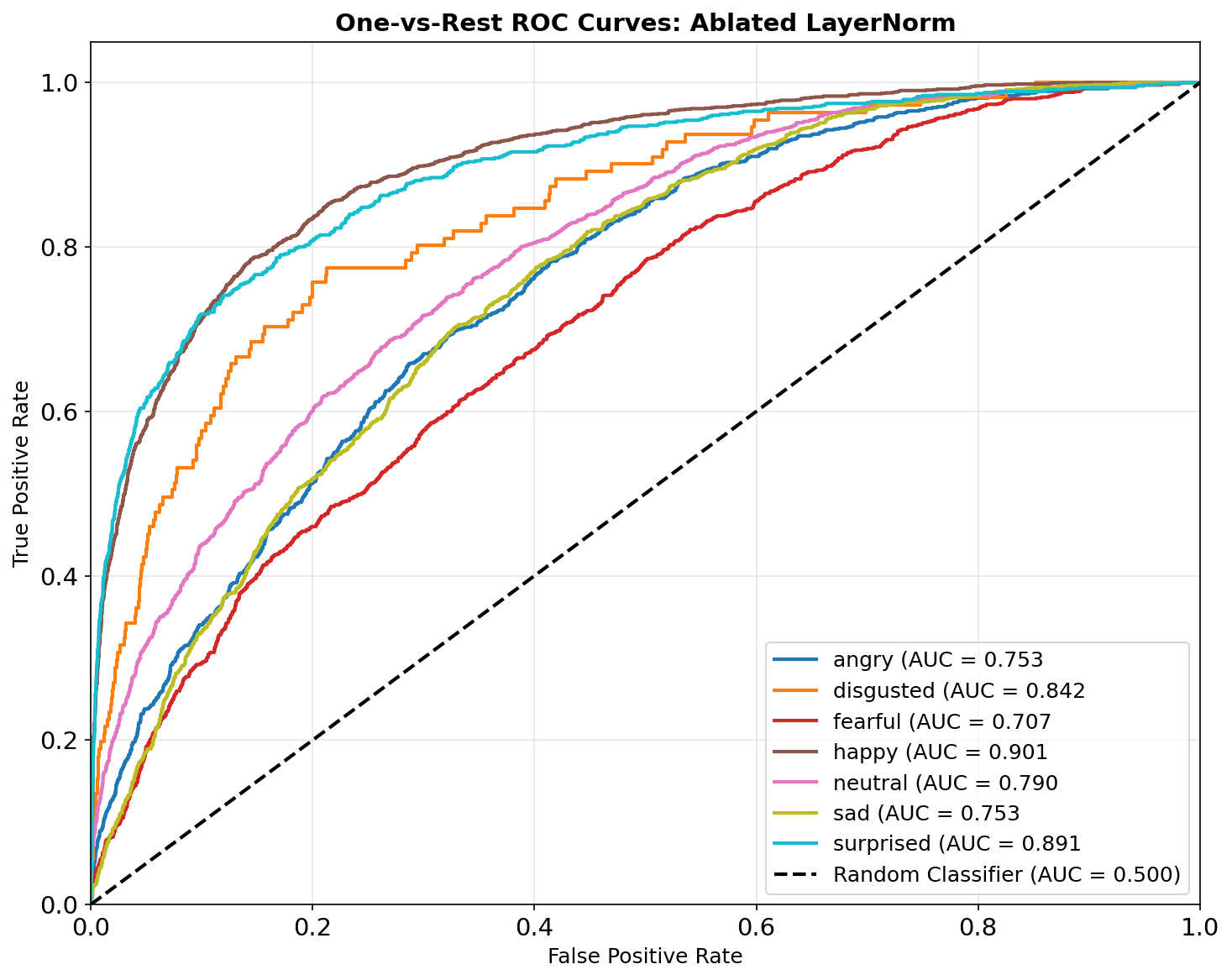}
\end{minipage}

\vspace{0.3em}

\noindent\begin{minipage}[t]{0.44\textwidth}
\includegraphics[height=0.24\textheight,width=\textwidth,keepaspectratio]{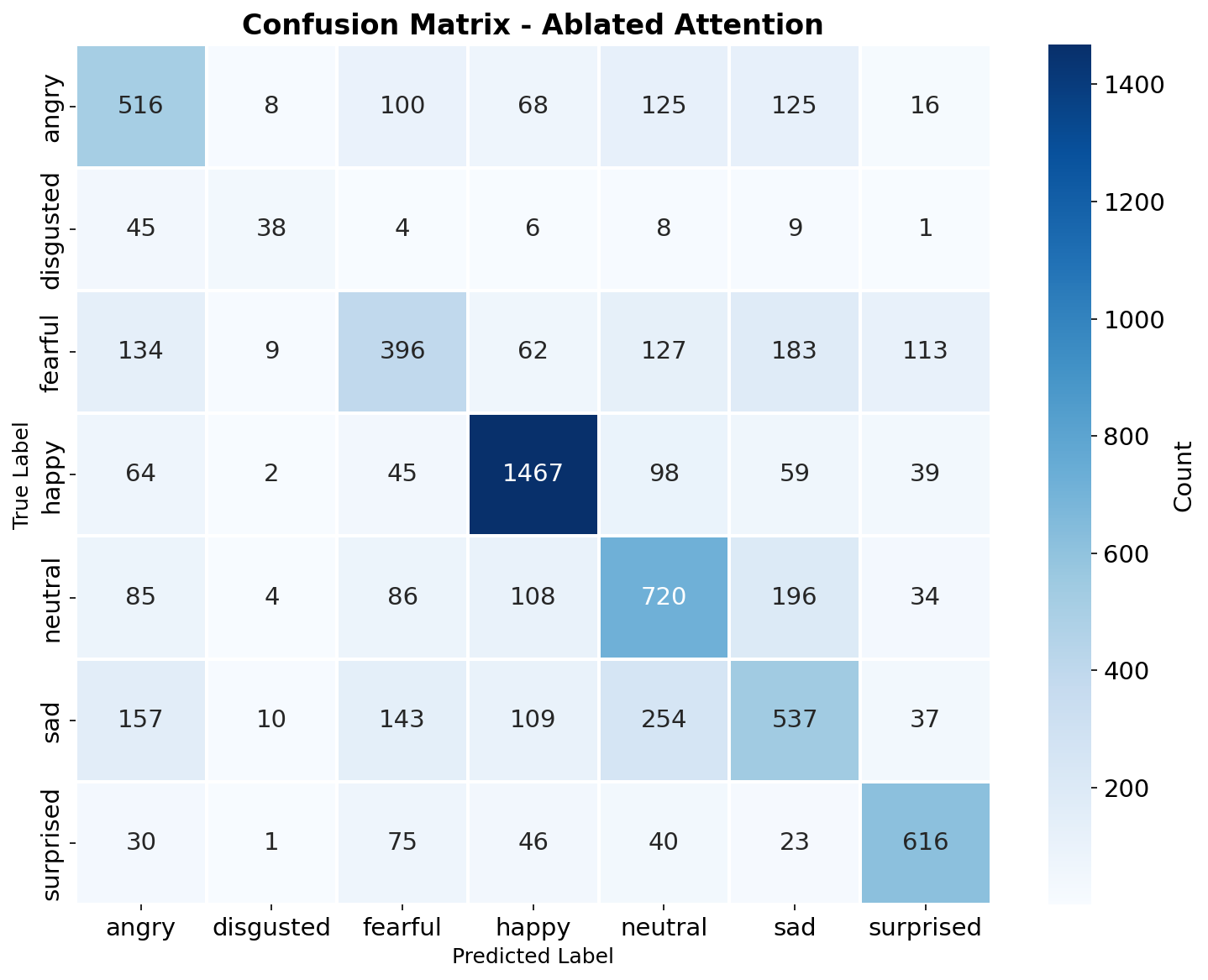}
\end{minipage}%
\hfill
\begin{minipage}[t]{0.44\textwidth}
\includegraphics[height=0.24\textheight,width=\textwidth,keepaspectratio]{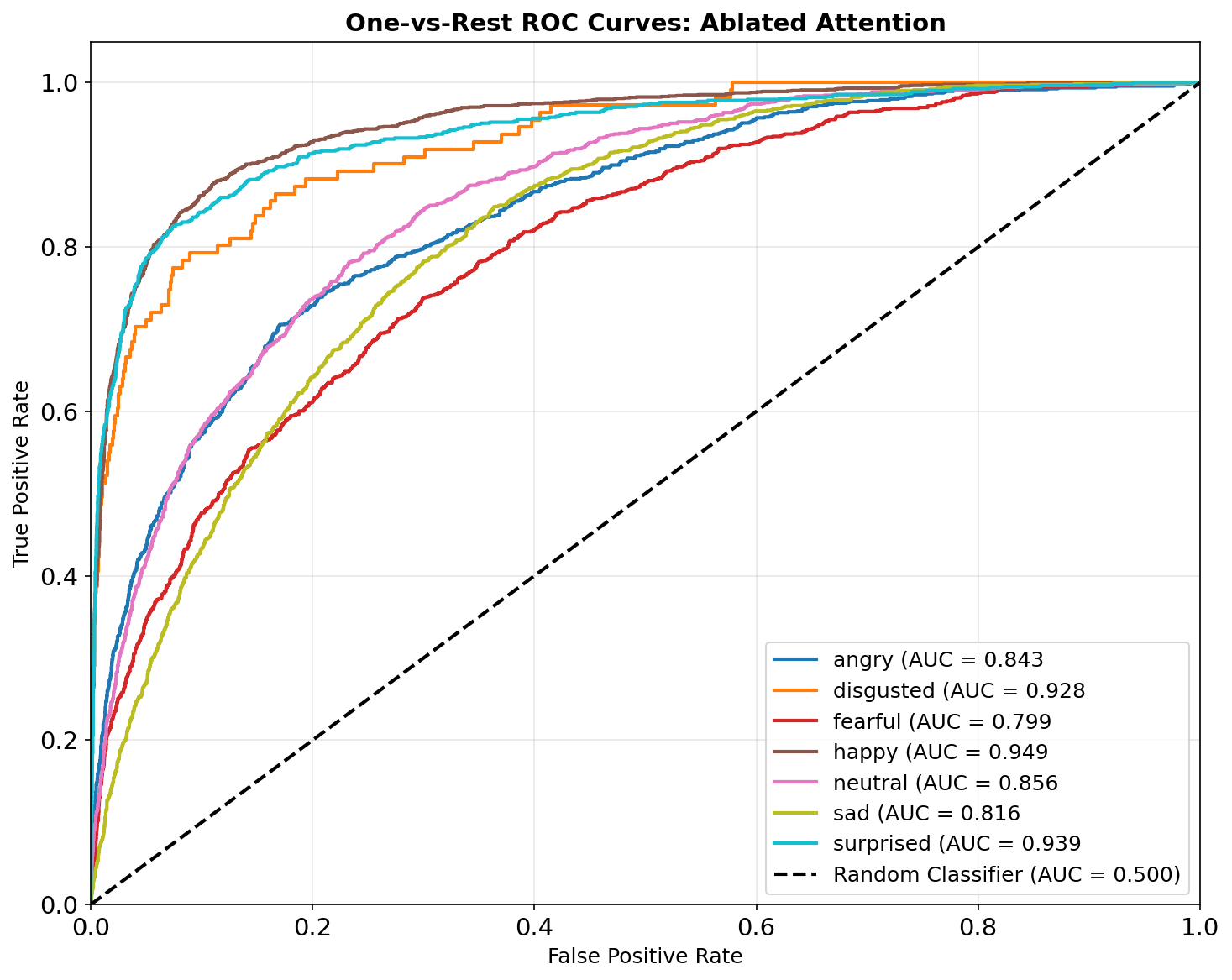}
\end{minipage}

\vspace{0.3em}

\end{figure*}
\end{document}